\documentclass[journal]{IEEEtran}
\ifCLASSINFOpdf
   \usepackage[pdftex]{graphicx}
\else
\fi
\usepackage{amsfonts,amssymb,amsmath}
\usepackage[ruled]{algorithm2e}
\usepackage{caption2}

\usepackage{color}
\begin{document}
\title{Deep Continuous Conditional Random Fields with Asymmetric Inter-object Constraints for Online Multi-object Tracking}

\author{Hui Zhou,
        Wanli Ouyang,
        Jian Cheng,
        Xiaogang Wang,
        and Hongsheng Li\thanks{Hui Zhou and Jian Cheng are with the School of Information and Communication Engineering at University of Electronic Science and Technology of China, Chengdu, China. Hongsheng Li and Xiaogang Wang are with the Department of Electronic Engineering at The Chinese University of Hong Kong, Hong Kong, China. Wanli Ouyang is with University of Sydney, Sydney, Australia. This work is done when Hui Zhou is a Research Assistant in the Department of Electronic Engineering at The Chinese University of Hong Kong. Hongsheng Li is the corresponding author (e-mail: hsli@ee.cuhk.edu.hk).}}% <-this % stops a space
% <-this % stops a space


\markboth{Journal of \LaTeX\ Class Files, DOI 10.1109/TCSVT.2018.2825679}%
{Shell \MakeLowercase{\textit{et al.}}: Bare Demo of IEEEtran.cls for IEEE Journals}

\IEEEpubid{\begin{minipage}{\textwidth}\ \\[12pt] \centering
  Copyright \copyright 20xx IEEE. Personal use is permitted, However, permission to use this material for any other purposes\\ 
  must be obtained from the IEEE by sending an email to pubs-permissions@ieee.org.
\end{minipage}} 


\maketitle

\begin{abstract}
Online Multi-Object Tracking (MOT) is a challenging problem and has many important applications including intelligence surveillance, robot navigation and autonomous driving. In existing MOT methods, individual object's movements and inter-object relations are mostly modeled separately and relations between them are still manually tuned. In addition, inter-object relations are mostly modeled in a symmetric way, which we argue is not an optimal setting. To tackle those difficulties, in this paper, we propose a Deep Continuous Conditional Random Field (DCCRF) for solving the online MOT problem in a track-by-detection framework. The DCCRF consists of unary and pairwise terms. The unary terms estimate tracked objects' displacements across time based on visual appearance information. They are modeled as deep Convolution Neural Networks, which are able to learn discriminative visual features for tracklet association. The asymmetric pairwise terms model inter-object relations in an asymmetric way, which encourages high-confidence tracklets to help correct errors of low-confidence tracklets and not to be affected by low-confidence ones much. The DCCRF is trained in an end-to-end manner for better adapting the influences of visual information as well as inter-object relations. Extensive experimental comparisons with state-of-the-arts as well as detailed component analysis of our proposed DCCRF on two public benchmarks demonstrate the effectiveness of our proposed MOT framework.
\end{abstract}

\begin{IEEEkeywords}
Multi-object tracking, Deep neural networks, Continuous Conditional Random Fields, Asymmetric pairwise terms.
\end{IEEEkeywords}

% \fi
\IEEEpeerreviewmaketitle

\section{Introduction}

\IEEEPARstart {R}{obust} tracking of multiple objects \cite{Luo2014Multiple} is a challenging problem in computer vision and acts as an important component of many real-world applications. It aims to reliably recover trajectories and maintain identities of objects of interest in an image sequence. State-of-the-art Multi-Object Tracking (MOT) methods \cite{xiang2015learning}, \cite{hong2016online} mostly utilize the tracking-by-detection strategy because of its robustness against tracking drift. Such a strategy generates per-frame object detection results from the image sequence and associates the detections into object trajectories. It is able to handle newly appearing objects and is robust to tracking drift. The tracking-by-detection methods can be categorized into offline and online methods. The offline methods  \cite{tang2016multi} use both detection results from past and future with some global optimization techniques for linking detections to generate object trajectories. The online methods, on the other hand, use only detection results up to the current time to incrementally generate object trajectories. Our proposed method focuses on online MOT, which is more suitable for real-time applications including autonomous driving and intelligent surveillance.

In MOT methods, the tracked objects usually show consistent or slowly varying appearance across time. Visual features of the objects are therefore important cues for associating detection boxes into tracklets. In recent years, deep learning techniques have shown great potential in learning discriminative visual features for single-object and multi-object tracking.
However, visual cues alone cannot guarantee robust tracking results. When tracked objects with similar appearances occlude or are close to each other, their trajectories might be wrongly associated to other objects. In addition, there also exist mis-detections or inaccurate detections by imperfect object detectors. Such difficulties escalate when the camera is hold by hand or fixed on a car. Each object moves according to its own movement pattern as well as the global camera motion. Solving such problems was explored by modeling interactions between tracked objects in the optimization model. For online MOT methods, there were investigations on modeling inter-object interactions with social force models \cite{pellegrini2009you, alahi2016social, leal2011everybody}, relative spatial and speed differences \cite{chen2016multiperson, zhang2014preserving, duan2012group}, and relative motion constraints \cite{hong2016online, yoon2015bayesian}.
Most of the previous methods model pairwise inter-object interactions in symmetric mathematical forms, i.e., pairs of objects influence each other with the same magnitude.

\IEEEpubidadjcol
However, such pairwise object interactions should be directional and modeled in an asymmetric form, while existing methods model such interactions in a symmetric way. For instance, large-size detection boxes are more likely to be noisy (if measured in actual pixels). Smaller boxes should influence larger boxes more than large ones to small ones because the smaller ones usually provide more accurate localization for objects. Similarly, high-confidence trajectories should influence low-confidence ones more and low-confidence ones should have minimal impact on the high-confidence ones. In this way, the more accurate detections or trajectories could help correct errors of the inaccurate ones and would not be affected by the inaccurate ones much. Moreover, in existing methods, individual object's movements and inter-object interactions are usually modeled separately. The relations between the two terms are mostly manually tuned and not effectively studied in a unified framework.
\IEEEpubidadjcol

To tackle the difficulties, we propose a Deep Continuous Conditional Random Field (DCCRF) with asymmetric inter-object constraints for solving the problem of online MOT. The DCCRF inputs a pair of consecutive images at time $t-1$ and time $t$, and tracked object's past trajectories up to time $t-1$. It estimates locations of the tracked objects at time $t$. The DCCRF optimizes an objective function with two terms, the unary terms, which estimate individual object's movement patterns, and the asymmetric pairwise terms, which model interactions between tracked objects. The unary terms are modeled by a deep Convolutional Neural Network (CNN), which is trained to estimate each individual object's displacement between time $t-1$ and time $t$ with each object's visual appearance. The asymmetric pairwise terms aim to tackle the problem caused by object occlusions, object mis-detections and global camera motion. For two neighboring tracked trajectories, the pairwise influence is different along each direction to let the high-confidence trajectory assists the low-confidence one more. Our proposed DCCRF utilizes mean-field approximation for inference and is trained in an end-to-end manner to estimate the optimal displacement for each tracked object. Based on such estimated object locations, a final visual-similarity CNN is proposed for generating the final detection association results.

\begin{figure*}[t]
  \begin{center}
    \includegraphics[width=1.0\linewidth]{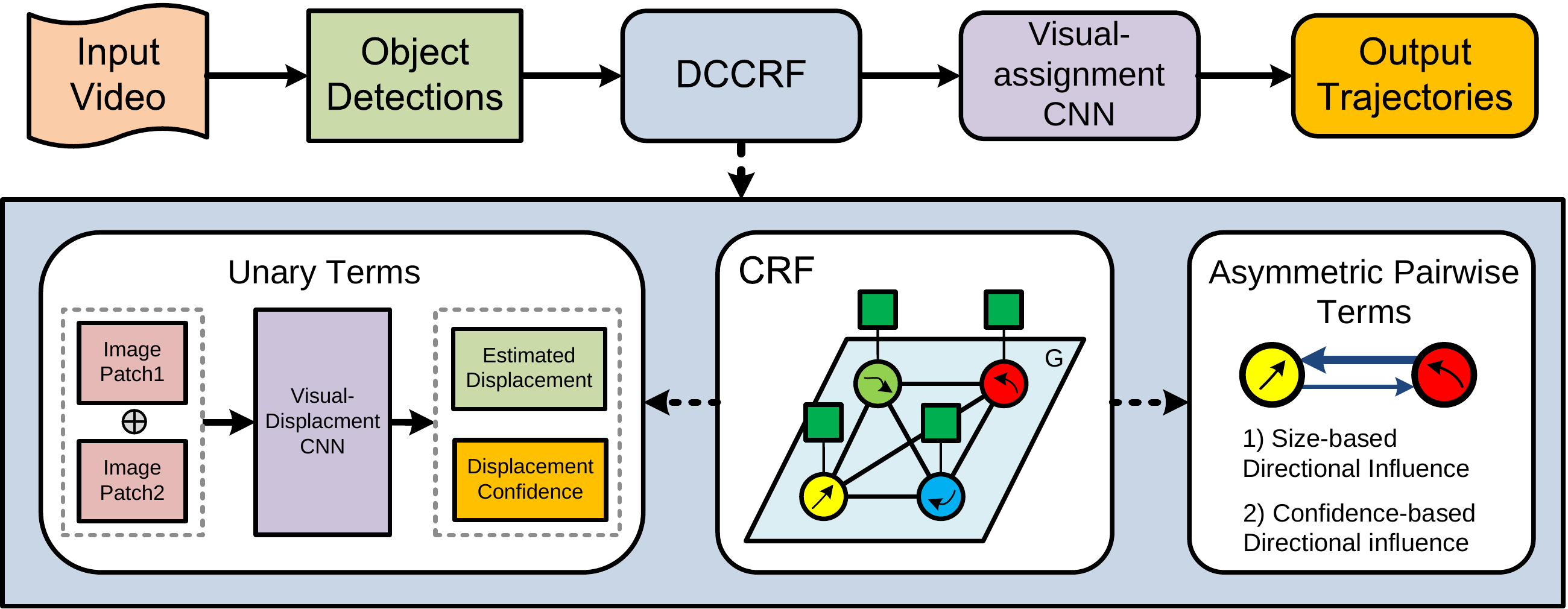}
  \end{center}
  \caption{Illustration of the overall multi-object tracking framework. The proposed Deep Continuous Conditional Random Field consists of unary terms and asymmetric pairwise terms (Section \ref{ssec:dccrf}). The unary terms are modeled by a visual-displacement CNN, which take pairs of object image patches as inputs and output the estimated object displacements between time $t-1$ and time $t$ (Section \ref{sssec:unary}). The asymmetric pairwise terms encourage to use high-confidence tracklets for correcting errors of low-confidence tracklets (Section \ref{sssec:asymmetric}). Size-based and confidence-based directional weighting functions are investigated.}
  \label{fig:overall}
\end{figure*}

The contribution of our proposed online MOT framework is two-fold. (1) A novel DCCRF model is proposed for solving the online MOT problem. Each object's individual movement patterns as well as inter-object interactions are studied in a unified framework and trained in an end-to-end manner. In this way, the unary terms and pairwise terms of our DCCRF can better adapt each other to achieve more accurate tracking performance.
(2) An asymmetric inter-object interaction term is proposed to model the directional influence between pairs of objects, which aims to correct errors of low-confidence trajectories while maintain the estimated displacements of the high-confidence ones.
Extensive experiments on two public datasets show the effectiveness of our proposed MOT framework.

\section{Related Work}
There are a large number of methods on solving the multi-object tracking problem. We focus on reviewing online MOT methods that utilize interactive constraints, as well as single-object and multi-object tracking algorithms with deep neural networks.

\textbf{Interaction models for MOT.}
%There exist a large number of MOT methods \cite{bae2014robust}, \cite{breitenstein2011online}, [] that directly used first or second order motion models to predict object locations during tracking. However, these methods did not work well on generally calibrated cameras and did not consider geometric layouts of the scenes.
Social force models were adopted in MOT methods \cite{pellegrini2009you, alahi2016social, leal2011everybody} to model pairwise interactions (attraction and repulsion) between objects. These methods required objects' 3D positions for modeling inter-object interactions, which were obtained by visual odometry.
%However, the interactions in both methods are defined in symmetric manner, i.e. pairs of objects influence each other with the same magnitude. The directional influence has not been yet much explored.

Grabner et al. \cite{grabner2010tracking} assumed that the relative positions between feature points and objects were more or less fixed over short-time intervals. Generalized Hough transform was therefore used to predict each target's location with the assist of supporter feature points.
%It was able to reduce tracking drift caused by object occlusion or its drastic appearance change.
Duan et al. \cite{duan2012group} proposed mutual relation models to describe the spatial relations between tracked objects and to handle occluded objects. Such constraints are learned by an online structured SVM. Zhang and  Maaten \cite{zhang2014preserving} incorporated spatial constraints between objects into an MOT framework to track objects with similar appearances.

The CRF algorithm \cite{zheng2015conditional} was used frequently in segmentation tasks to model the relationship between different pixels in the spatial-domain. There were also many works that modeled the multi-object tracking problem with CRF models. Yang and Nevatia \cite{yang2012online} proposed an online-learned CRF model for MOT, and assumed linear and smooth motion of the objects to associate past and future tracklets. Andriyenko et al. \cite{andriyenko2012discrete} modeled multi-object tracking as optimizing discrete and continuous CRF models. A continuous CRF was used for enforcing motion smoothness, and a discrete CRF with a temporal interaction pairwise term was optimized for data association. Milan et al. \cite{Milan_2013_CVPR} designed new CRF potenials for modeling spatio-temporal constraints between pairs of trajectories to tackle detection and trajectory-level occlusions.

\textbf{Deep learning based object tracking.} Most existing deep learning based tracking methods focused on single object tracking, because deep neural networks were able to learn powerful visual features for distinguishing the tracked objects from the background and other similar objects. Early single-object tracking methods \cite{li2014robust}, \cite{wang2013learning} with deep learning focused on learning discriminative appearance features for online training. However, due to the large learning capacitity of deep neural networks, it is easy to overfit the data. \cite{hong2015online}, \cite{Wang_2015_ICCV} pretrained deep convolutional neural networks on large-scale image dataset to learn discriminative visual features, and updated the classifier online with new training samples. More recently, methods that did not require model updating were proposed. Tao et al. \cite{tao2016siamese} utilized Siamese CNNs to determine visual similarities between image pacthes for tracking. Bertinetto et al. \cite{bertinetto2016fully} changed the network into a fully convolutional setting and achieved real-time running speed.

Recently, deep models have been applied to multi-object tracking. Milan et al. \cite{milan2017online} proposed an online MOT framework with two RNNs. One RNN was used for state (object locations, motions, etc.) prediction and update, and the other for associating objects across time. However, this method did not utilize any visual feature and relied solely on spatial locations of the detection results. \cite{bae2017confidence}, \cite{leal2016learning} replaced the hand-crafted features (e.g., color histograms) with the learned features between image patches by a Siamese CNN, which increases the discriminative ability. However, those methods focused on modeling individual object's movement patterns with deep learning. Inter-object relations were not integrated into deep neural networks.

\section{Method}

The overall framework of our proposed MOT method is illustrated in Fig. \ref{fig:overall}. We propose a Deep Continuous Conditional Random Field (DCCRF) model for solving the online MOT problem. At each time $t$, the framework takes past tracklets up to time $t-1$ and detection boxes at time $t$ as inputs, and generates new tracklets up to time $t$. At each time $t$, new tracklets are also initialized and current tracklets are terminated if tracked objects disappear from the scene.

The core components of the proposed DCCRF consist of unary terms and asymmetric pairwise terms. The unary terms of our DCCRF are modeled by a deep CNN that estimates the individual tracked object's displacements between consecutive times $t-1$ and $t$. The asymmetric pariwise terms aim to model inter-object interactions, which consider differences of speeds, visual-confidence, and object sizes between neighboring objects. Unlike interaction terms in existing MOT methods, which treat inter-object interactions in a symmetric way, asymmetric relationship terms are proposed in our DCCRF. For pairs of tracklets in our DCCRF model, the proposed asymmetric pairwise term models the two directions differently, so that high-confidence trajectories with small-size detection boxes can help correct errors of low-confidence trajectories with large-size detection boxes. Based on the estimated object displacements by DCCRF, we adopt a visual-similarity CNN and Hungarian algorithm to obtain the final tracklet-detection associations.

\subsection{Deep Continuous Conditional Random Field (DCCRF)}
\label{ssec:dccrf}

The proposed DCCRF takes object trajectories up to time $t-1$ and video frame at time $t$ as inputs, and outputs each tracked object's displacement between time $t-1$ and time $t$.
Let ${\bf r}$ represents a random field defined over a set of variables $\{r_1, r_2, \cdots, r_n\}$, where each of the $n$ variables represents the visual and motion information of an object tracklet.
Let ${\bf d}$ represents another random field defined over variables $\{d_1, d_2, \cdots, d_n\}$, where each variable represents the displacement of an object between time $t-1$ and time $t$. The domain of each variable is the two-dimensional space $\mathbb{R}^2$, denoting the $x$- and $y$-dimensional displacements of tracked objects. Let $I$ represents the new video frame at time $t$.

The goal of our conditional random field $({\bf r}, {\bf d})$ is to maximize the following conditional distribution,
\begin{align}
P({\bf d}|{\bf r}, I) = \frac{1}{Z({\bf t})}\exp\left( -E({\bf d}, {\bf r}, I) \right),
\end{align}
where $E({\bf d}, {\bf r}, I)$ represents the Gibbs energy and $Z({\bf t})=\int_{\bf r} \exp(-E({\bf d}, {\bf r})) d{\bf r}$ is the partition function. Maximizing the conditional distribution w.r.t. ${\bf d}$ is equivalent to minimizing the Gibbs energy function,
\begin{align}
E({\bf d}, {\bf r}, I) = \sum_{i=1}^n \phi(d_i, r_i, I) + \sum_{i,j} \psi(d_i, d_j, r_i, r_j, I),
\label{eqn:energy}
\end{align}
where $\phi(d_i, r_i, I)$ and $\psi(d_i, d_j, r_i, r_j)$ are the unary terms and pairwise terms.

After the displacements ${\bf d}$ of tracked objects between time $t-1$ and time $t$ are obtained, individual object's estimated locations at time $t$ can be easily calculated for associating tracklets and detection boxes to generate tracklets up to time $t$. Such displacements are then iteratively calculated for the following time frames. Without loss of generality, we only discuss the approach for optimizing object displacements between time $t-1$ and time $t$ in this section.

%%Here we introduce the basic theory of CRF modeling and describe the details of our deep continuous CRF model.
%We denote $x \in \mathcal{X}$ as input and $y \in \mathcal{Y}$ as output. The energy function is defined by $E(y,x;\theta)$ which models the relationship of the input-output pair. All network parameters $\theta$ are learned. The conditional distribution is formulated as follows:
%\begin{equation}
%  P(y|x) = \frac{1}{Z(x)}exp\{-E(y,x)\}
%  \label{Equ:B_CRF}
%\end{equation}
%where $Z(x)=\int exp\{-E(y,x)\}d_{x}$ is the partition function. Many methods(e.g. mean field, belief propagation) can be found in the previous literatures[][] to fit or approximate the distribution.

%Given a sets of tracklets $S=\{T_{1}, T_{2},...,T_{n}\}$ and a sets of predictions $P=\{D_{1}, D_{2},...,D_{n}\}$ obtained from the visual motion CNN. A graph $G=(N,E)$ is created for formulate the interaction among the different tracklets in $S$. Each node is defined as a tracklet. We introduce a symbol $l_{i}$ to represent the optimal displacement. The full energy function is as follows:
%\begin{equation}
%  E(L)=\sum_{i}U(l_{i}|S,D) + \sum_{ij}B(l_{i},l_{j}|S,D)
%  \label{Equ:overall}
%\end{equation}
%where $U$ is unary term and $B$ is pairwise term.

%\IEEEpubidadjcol

\subsubsection{\bf{Unary terms}}
\label{sssec:unary}

For the $i$th object tracklet, the unary term $\phi(d_i, r_i, I)$ of our DCCRF model is defined as
\begin{align}
\phi(d_i, r_i, I) = w_{i,1} \left(d_i - f_d(r_i, I)\right)^2.
\label{eqn:unary}
\end{align}
This term penalizes the quadratic deviations between the final output displacement $d_i$ and the estimated displacement by a visual displacement estimation function $f_d$. $w_{i,1}$ is an online adaptive parameter for the $i$th object that controls to trust more the estimated displacement based on the $i$th object's visual cues (the unary terms) or based on inter-object relations (the pairwise terms). Intuitively, when the visual displacement estimator $f_d$ has higher confidence on its estimated displacement, $w_{i,1}$ should be larger to bias the final output $d_i$ towards the visually inferred displacements. On the other hand, when $f_d$ has lower confidence on its estimation, due to object occlusion or appearing of similar objects, $w_{i,1}$ should be smaller to let the final displacement $d_i$ be mainly inferred by inter-object constraints.

In our framework, the visual displacement estimation function $f_d$ is modeled as a deep Convolution Neural Network (CNN) that utilizes only the tracked objects' visual information for estimating its location displacement between time $t-1$ and time $t$. For each tracked object $r_i$, our visual-displacement CNN takes a pair of images patched from frames $t-1$ and $t$ as inputs, and outputs the object's inferred displacement. A network structure similar to ResNet-101 \cite{He_2016_CVPR} (except for the topmost layer) is adopted for our visual-displacement CNN.
\begin{figure}
  \begin{center}
    \includegraphics[width=1.0\linewidth]{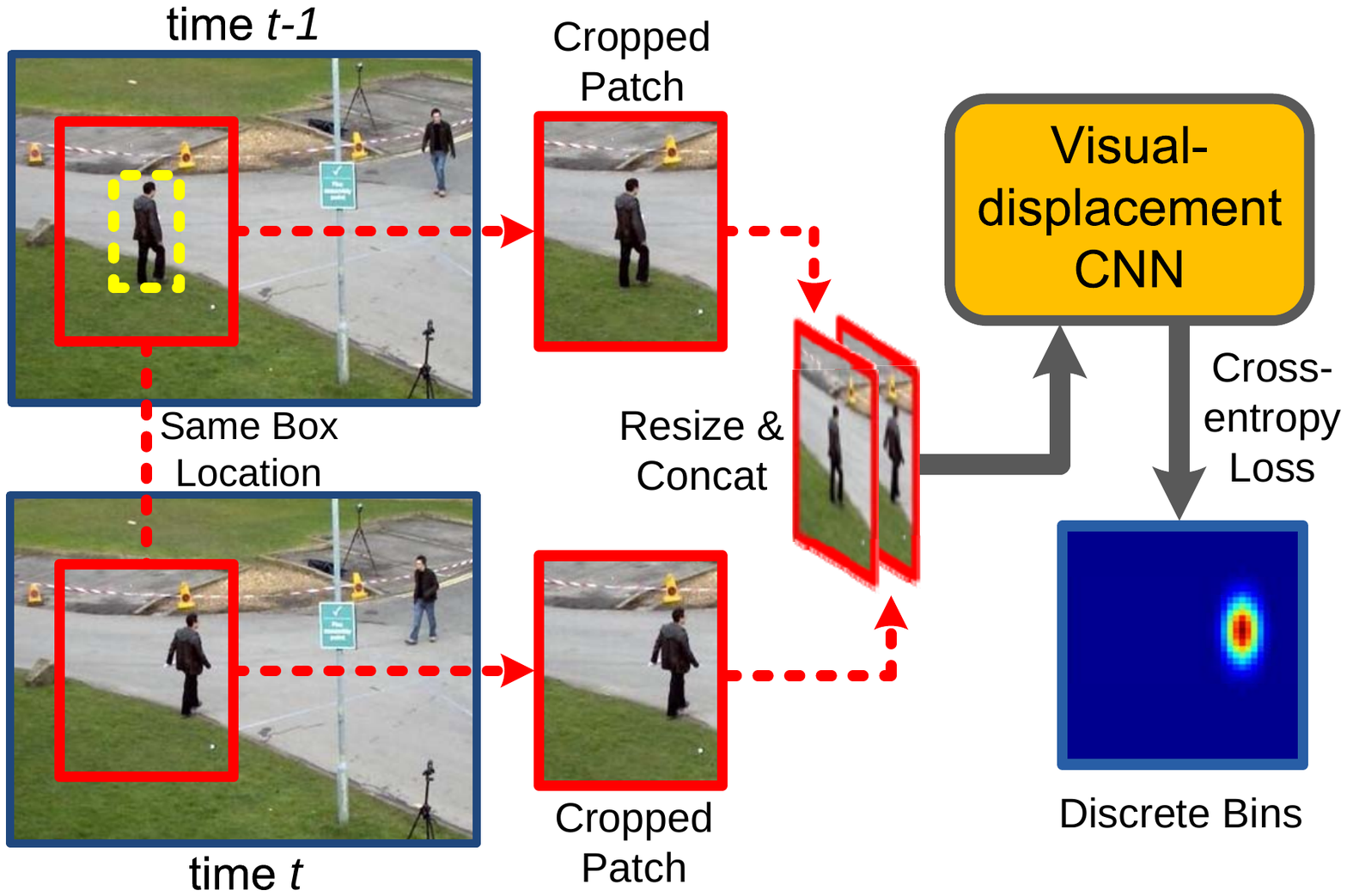}
  \end{center}
  \caption{Illustration of the visual-displacement CNN for modeling the unary terms. Two image patches are cropped from the same box location centered at the object location at time $t-1$ as inputs. The visual-displacement CNN estimates the confidences of discrete object displacements and is trained with cross-entropy loss.}
  \label{fig:unary}
\end{figure}
The network inputs and outputs are illustrate in Fig. \ref{fig:unary}. For the inputs, given currently tracked object $r_i$'s bounding box location $b_i$ at time $t-1$, a larger bounding box $\bar{b}_i$ centered at $b_i$ is first created. Two image patches are cropped at the same spatial location $\bar{b}_i$ but from different frames at time $t-1$ and time $t$. They are then concatenated along the channel dimension to serve as the inputs for our visual-displacement CNN. The reasons for using a larger bounding box $\bar{b}_i$ instead of the original box $b_i$ are to tolerate large possible displacement between the two consecutive frames and also to incorporate more visual contextual information of the object for more accurate displacement estimation. After training with thousands of such pairs, the visual-displacement CNN is able to capture important visual cues from image-patch pairs to infer object displacements between time $t-1$ and time $t$.

For the CNN outputs, instead of directly estimating objects' two dimensional $x$- and $y$-dimensional displacements, we discretize possible 2D continuous displacements into a 2D discrete grid $\{p_i^1, p_i^2, \cdots, p_i^m\}$ (bottom-right part in Fig. \ref{fig:unary}), where $p_i^k \in \mathbb{R}^2$ represents the displacement corresponding to the $k$th bin of the $i$th object. The visual-displacement CNN is trained to output confidence scores $c_i^k$ for the displacement bins $p_i^k$ with a softmax function. The cross-entropy loss is therefore used to train the CNN, and the final estimated displacement for the tracked object $r_i$ is calculated as the weighted average of all possible displacements $\sum_{k=1}^m c_i^k p_i^k$, where $\sum_{k=1}^m c_i^k=1$.
In practice, we discretize displacements into $m=20\times20$ bins, which is a good trade-off between discretization accuracy and robustness. Note that there are existing tracking methods \cite{bertinetto2016fully}, \cite{held2016learning} that also utilize pairs of image patches as inputs to directly estimate object displacements. However, in our method, we propose to use cross-entropy loss for estimating displacements and find that its result achieves more accurate and robust displacement estimations in our experiments.
More importantly, it provides displacement confidence scores $\{c_i^1,\cdots, c_i^m\}$ for calculating the adaptive parameter $w_{i,1}$ in Eq. (\ref{eqn:unary}) to weight the unary and pairwise terms.

The confidence weight $w_{i,1}$ is obtained by the following equation,
\begin{align}
w_{i,1} = \sigma\left(a_1 \max({\bf c}_i) + b_1\right),
\end{align}
where $\sigma$ is the sigmoid function constraining the range of $w_{i,1}$ being between 0 and 1, $\max({\bf c}_i)$ obtains the maximal confidence of ${\bf c}_i=\{c_i^1,c_i^2,\cdots, c_i^m\}$, and $a_1$ and $b_1$ are learnable scalar parameters. In our experiments, the learned parameter $a_1$ is generally positive after training, which denotes that, if the visual-displacement CNN is more confident about its displacement estimations, the value of $w_{i,1}$ is larger and the final output displacement $d_i$ can be more biased towards the visually inferred displacement $f_d(r_i, I)$. Otherwise, the final displacement $d_i$ can be biased to be inferred by inter-object constraints.

If the energy function $E$ in Eq. (\ref{eqn:energy}) consists of only the unary terms $\phi(d_i, r_i, I)$, the final output displacement $d_i$ can be solely dependent on each tracked object's visual information without considering inter-object constraints.

\subsubsection{\bf{Asymmetric pairwise terms}}
\label{sssec:asymmetric}

The pairwise terms in Eq. (\ref{eqn:energy}) are utilized to model asymmetric inter-object relations between object tracklets for regularizing the final displacement results ${\bf d}$.
%Generally, the high-confidence displacement estimations should be used to correct errors of low-confidence ones. On the contrary, the low-confidence estimations should not influence the high-confidence ones.
To handle global camera motion, we assume that from time $t-1$ to time $t$, the speed differences between two tracked objects should be maintained, i.e.,
\begin{align}
\psi(d_i, d_j, r_i, r_j, I) = (1-w_{i,1}) \sum_k w_{ij,2}^{(k)} \left(\Delta d_{ij} - \Delta s_{ij} \right)^2,
\label{eq:pairwise}
\end{align}
where $\Delta d_{ij}= d_i-d_j$ is the displacement (which can be viewed as speed) difference between objects $i$ and $j$ at time $t$, $\Delta s_{ij} = s_i - s_j$ is the speed difference at the previous time $t-1$, and $w_{ij,2}^{(k)}$ are a series of weighting functions (two in our experiments) that control the directional influences between the pair of objects,

\begin{figure}[!t]
  \begin{center}
    \includegraphics[width=1.0\linewidth]{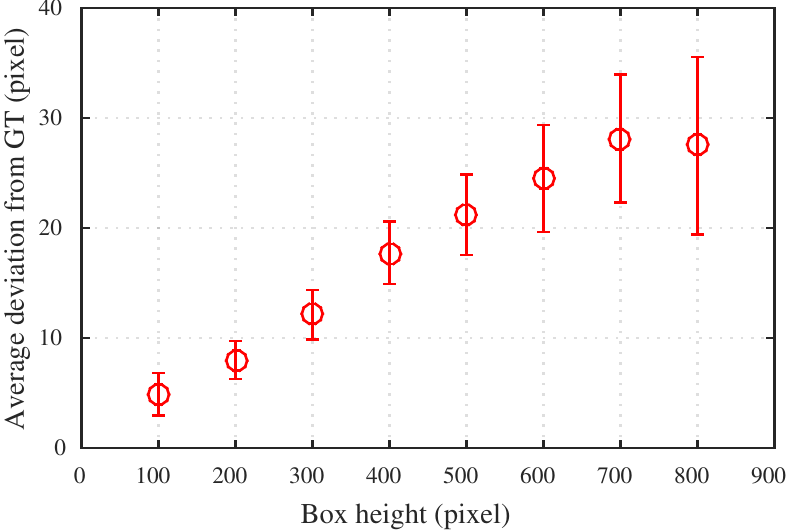}
  \end{center}
  \caption{The average deviation of detection boxes from their ground-truth locations is approximately proportional to the detection box size. The statistics are calculated from the 2DMOT16 training set \cite{milan2016mot16} where the detection boxes are provided by the dataset.}
  \label{fig:box_noise}
\end{figure}

\begin{figure}[!t]
	\centering
	\begin{tabular}{c@{\hspace{-2mm}}c@{\hspace{2mm}}c}
		&
		\includegraphics[scale=0.55]{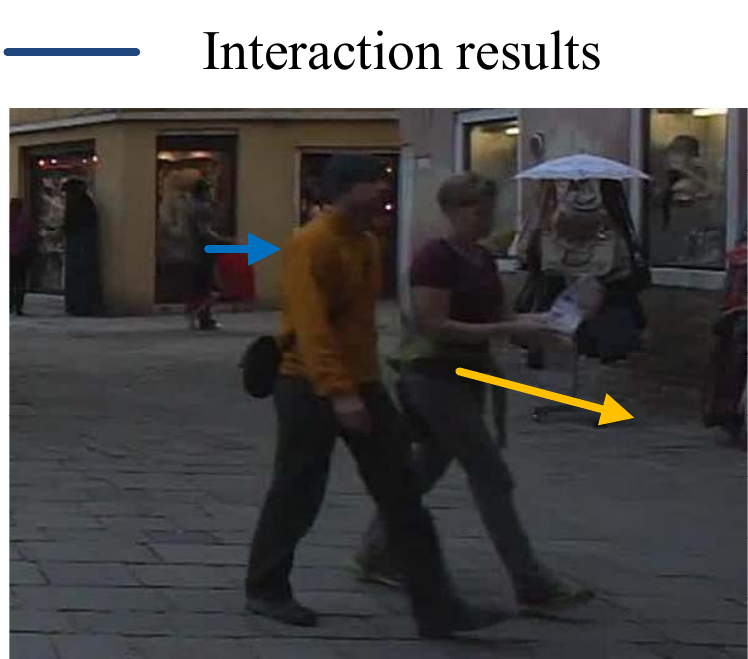}&
		\includegraphics[scale=0.55]{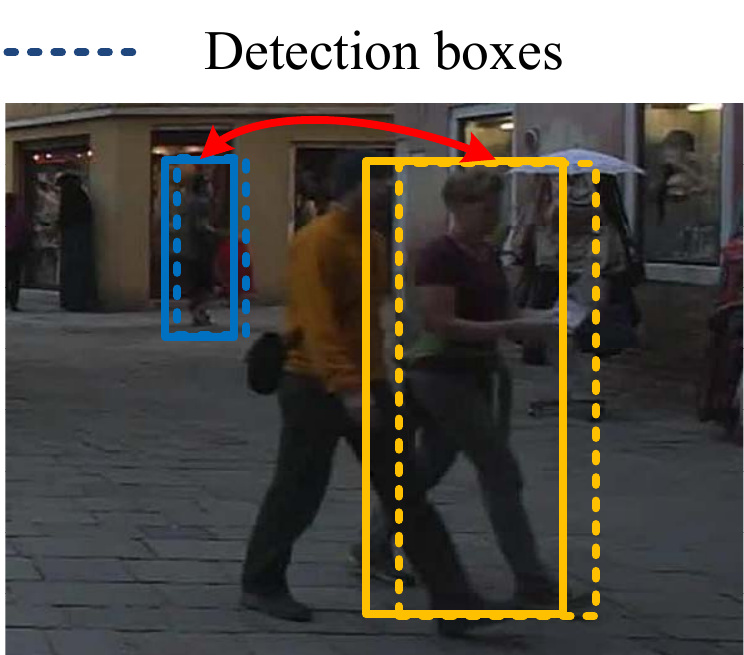}\\
		& (a) Time $t-1$ & (b) Time $t$ \\
		&                & Symmetric influence\\
		&
		\includegraphics[scale=0.55]{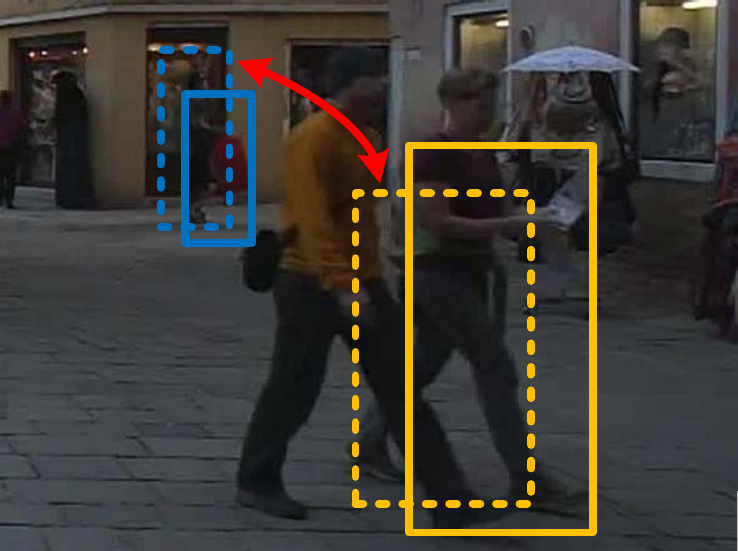}&
		\includegraphics[scale=0.55]{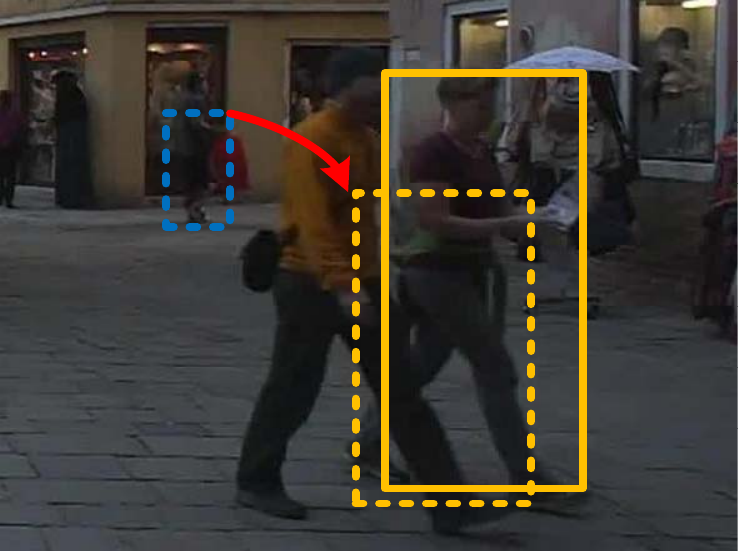}\\
		& (c) Time $t$ & (d) Time $t$ \\
		& Symmetric influence    & {\small Asymmetric small-to-large influence } \\
	\end{tabular}
  \caption{Illustration of symmetric and asymmetric inter-object constraints. (a) Two tracked objects at time $t-1$ with their estimated speeds (denoted by arrows). (b) Tracked objects at time $t$. Symmetric inter-object constraints work well when there are little detection noise for all detection boxes. (c) When there is localization noise for the large-size detection box, symmetric inter-object constraints are likely to deteriorate the tracking of the small-size object. (d) Asymmetric small-to-large-size inter-object constraints are more robust than symmetric inter-object constraints, even when there is localization noise for the small-size detection box.}
  \label{fig:intro}
\end{figure}

For better modeling inter-object relations, two important observations are made to define the asymmetric weighting functions $w_{ij,2}^{(k)}$. 1) For detection boxes, in terms of localization accuracy, larger object detection boxes are more likely to be noisy, while smaller ones tend to be more stable (as shown in Fig. \ref{fig:box_noise}). This is because the displacements of both large and small detection boxes are all recorded in pixels in our tracking frameworks. Noisy large detection boxes would significantly influence the displacement estimation for other boxes. This problem is illustrated in Fig. \ref{fig:intro}. The two targets in Fig. \ref{fig:intro}(a) have accurate locations and speeds which can be used to build inter-object constraints at time $t-1$. 
%and then we will introduce one successful example and one failing example to show limitations of symmetric interaction in existing methods. 
When the detector outputs roughly accurate bounding boxes for both targets at time $t$, symmetric inter-object constraints could well refine the objects' locations (see Fig. \ref{fig:intro}(b)). However, since the larger-size detection boxes are more likely to be noisy, using the symmetric inter-object constraints would significantly affect tracking results of the small-size objects (see Fig. \ref{fig:intro}(c)). In contrast, small-size objects have smaller localization errors and could better infer larger-size objects' locations. Asymmetric small-to-large-size inter-object constraints are robust, even when the smaller-size detection box is noisy(see Fig. \ref{fig:intro}(d)). Therefore, between a pair of tracked objects, the one with smaller detection box should have more influence to infer the displacement of the ones with larger detection box, and the object with a larger box should have less chance to deteriorate the displacement estimation of the smaller one.
%motion constraint at time $t-1$ can be used by tracklets refinement at time $t$ (see Fig. \ref{fig:intro}(b)). However, the large-size detection box is more likely to be noisy (see Fig. \ref{fig:intro}(c)) and deteriorates the tracking of the small-size object when we use symmetric inter-object constraints. Asymmetric inter-object constraints don't have these shortcomings, even if the small-size box is noisy (see Fig. \ref{fig:intro}(d)). Therefore, between a pair of tracked objects, the one with smaller detection box should have more influence to infer the displacement of the ones with larger detection box, and the object with a larger box should have less chance to deteriorate the displacement estimation of the smaller one.}
2) If our above mentioned visual-displacement CNN has high confidence for an object's displacement, this object's visually inferred displacement should be used more to infer other objects' displacements. On the other hand, the objects with low confidences on their visually inferred displacements should not affect other objects with high-confidence displacements.
Based on the two observations, we model the weighting function $w_{ij,2}^{(k)}$ by a product of a size-based weighting function and a confidence-based weighting function between a pair of tracked objects as
\begin{align}
w_{ij,2}^{(k)} = & \sigma (a_{21}^{(k)} \log(area_i/area_j) + b_{21}^{(k)})) \times\nonumber \\
& \sigma(a_{22}^{(k)} (\max({\bf c}_i) - \max({\bf c}_j)) + b_{22}^{(k)})
\end{align}
where $\sigma$ denotes the sigmoid function, $area_i$ denotes the size of the $i$th tracked object at time $t-1$, $\max ({\bf c}_i)$ obtains the maximal displacement confidence from $\{ c_i^1, c_i^2, \cdots, c_i^m \}$ by our proposed visual-displacement CNN, and $a_{21}^{(k)}$, $b_{21}^{(k)}$, $a_{22}^{(k)}$, $b_{22}^{(k)}$ are learnable scalar parameters. In our DCCRF, these parameters can be learned by back-propagation algorithm with mean-field approximation. If we use the mean-field approximation for DCCRF inference, the influence from object $r_i$ to $r_j$ and that from $r_j$ and $r_i$ are different (see next subsection for details). After training, we see that $a_{21}^{(k)} > 0$ and $a_{22}^{(k)} < 0$, which means that smaller $area_i/area_j$ and larger $\max({\bf c}_i) - \max({\bf c}_j)$ lead to greater weights. It validates our above mentioned observations that objects with smaller sizes and greater visual-displacement confidences should have greater influences to other objects, but not the other around.

\begin{figure}
	\centering
	\begin{tabular}{c@{\hspace{-1.5mm}}c@{\hspace{1mm}}c}
		&
		\includegraphics[height=32.5mm]{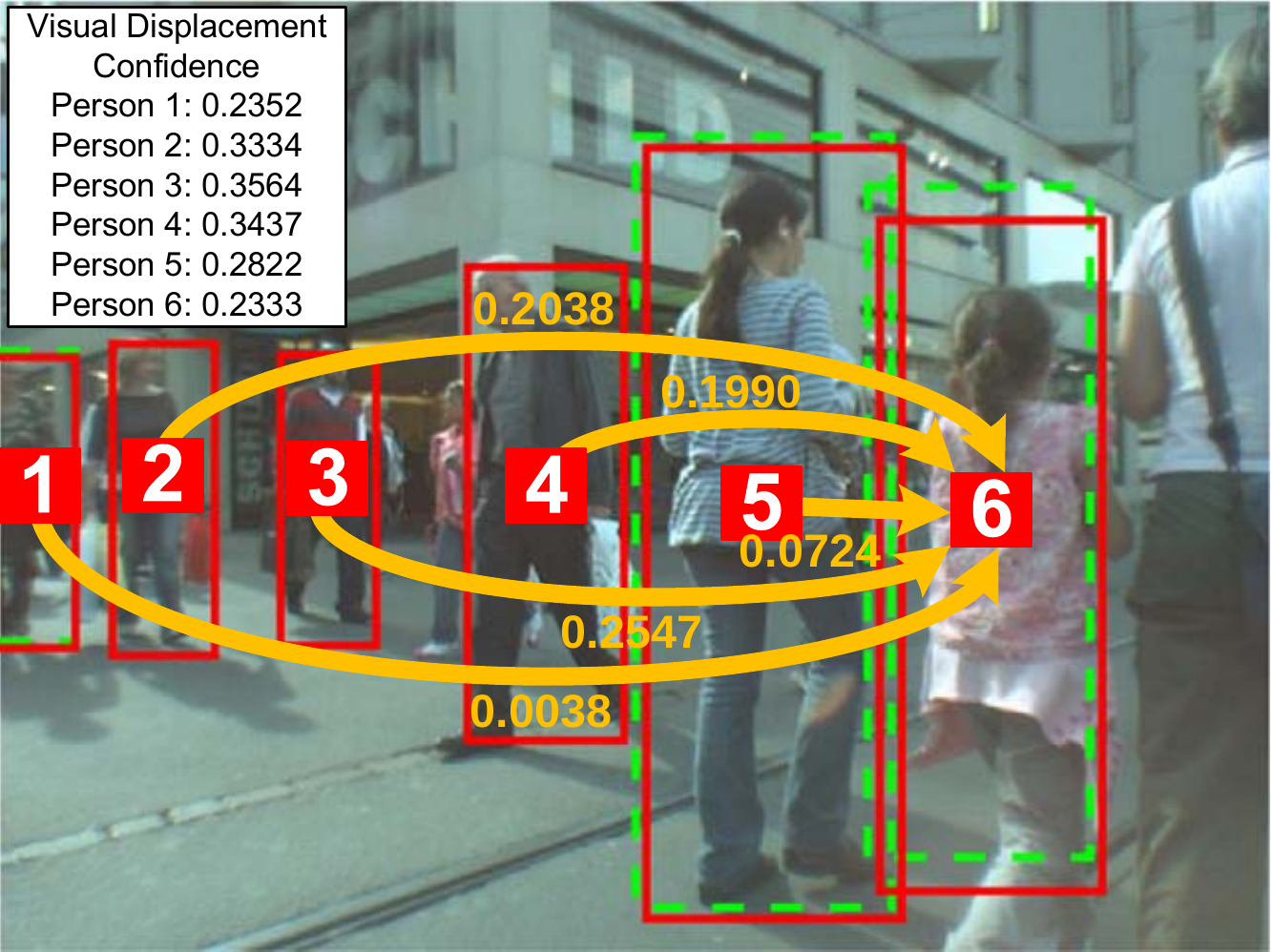}&
		\includegraphics[height=32.5mm]{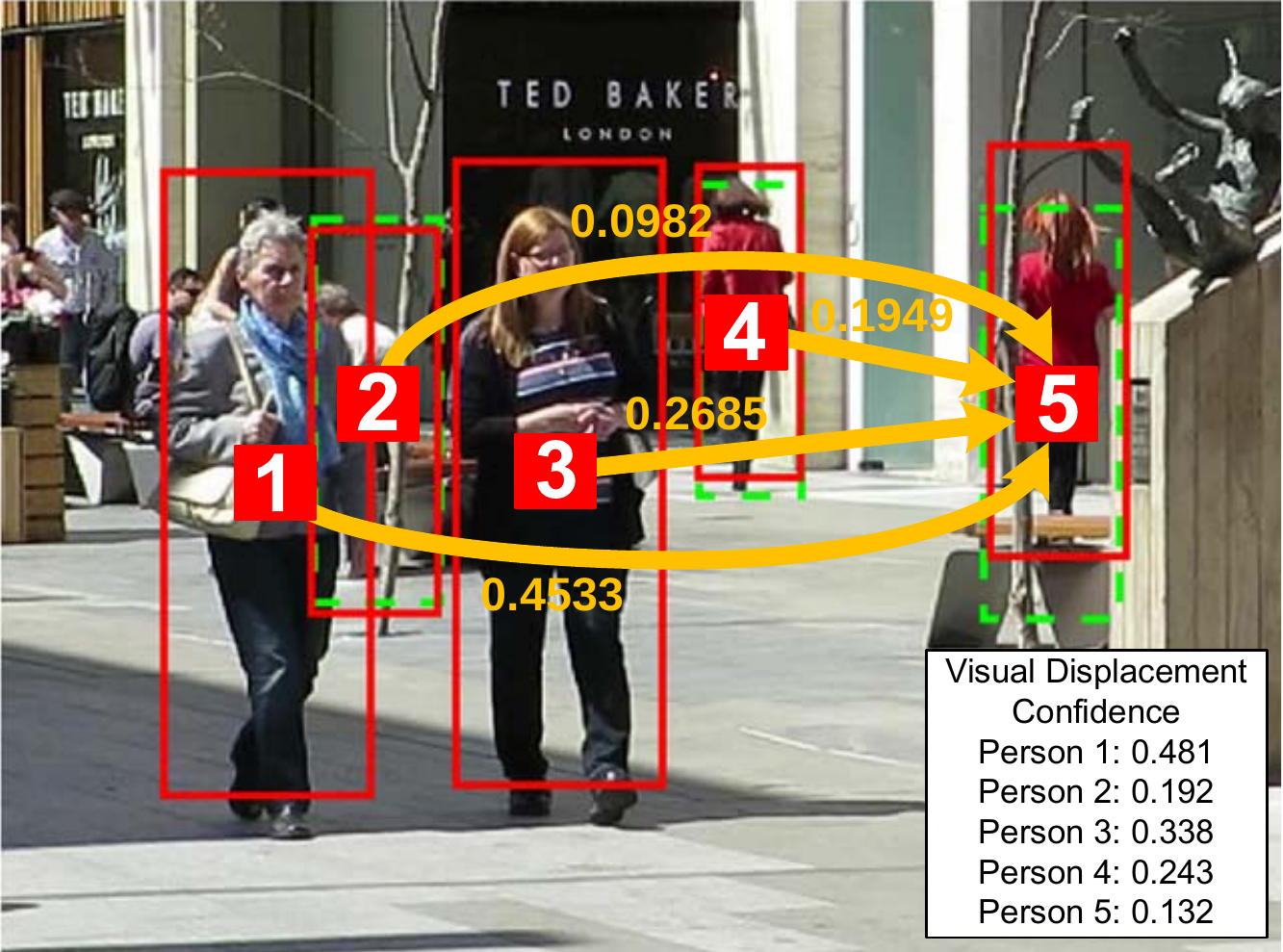}\\
		&(a)&(b)
	\end{tabular}
  \caption{Example values of asymmetric weighting function $w^{(k)}_{ij}$ between tracked objects of different sizes and confidences. Green dashed rectangles denote estimated object locations by the unary terms (visual-displacement CNN) only. Red rectangles denote estimated object locations by both unary and pairwise terms. Orange arrows and numbers denote the weighting function values from one object to the other. (a) Small-size objects (objects 2-4) help correct errors of large-size object (object 6) with higher diretional weighting function values. (b) Objects with higher visual-displacement confidences (objects 1, 3, 4) help correct errors of the object (object 5) with lower visual-displacement confidences.}
  \label{fig:pairwise}
\end{figure}

In Fig. \ref{fig:pairwise}, we show example values of one learned weighting function $w_{ij,2}^{(k)}$. In Fig. \ref{fig:pairwise}(a), compared with object 6, objects 2-4 are of smaller sizes and also higher visual confidences. With the directional weighting functions, they have greater influence to correct errors of tracking object 6 (red vs. green rectangles of object 6) and are not affected much by the erroneous estimation of object 6. Similar directional weighting function values can be found in Fig. \ref{fig:pairwise}(b), where objects 1, 3, 4 with high visual-displacement confidences are able to correct tracking errors of object 5 with low visual-displacement confidence.

\subsubsection{\bf{Inference}}

For our unary terms, we utilize forward propagation of the visual-displacement CNN for calculating objects' estimated displacements and displacement confidences $\{c_i^1, c_i^2, \cdots, c_i^m\}$. After the unary term inference, the overall maximum posterior marginal inference is achieved by mean-field approximation. This approximation yields an iterative message passing for approximate inference. Our unary terms and pairwise terms are both of quadratic form. The energy function is convex and the optimal displacement is obtained as the mean value of the energy function,
\begin{equation}
  d_{i} \longleftarrow \frac{w_{i,1}f_d(r_i, I)+(1-w_{i,1})\sum_{i \neq j} \sum_{k} w_{ij, 2}^{(k)}(d_j-\Delta s_{ij})}{w_{i,1}+(1-w_{i,1})\sum_{i \neq j} \sum_{k} w_{ij,2}^{(k)}}.
  \label{Equ:infer_param}
\end{equation}
In each iteration, the node $i$ receives messages from all other objects to update its displacement estimation. The mean-field approximation is usually converged in 5-10 iterations. The above displacement update equation clearly shows the differences between the messages transmitted from $i$ to $j$ and that from object $j$ to $i$ because of the asymmetric weighting functions $w_{ij,2}^{(k)}$. For a pair of objects, $w_{ij,2}^{(k)}$ and $w_{ji,2}^{(k)}$ are generally different. Even if $w_{i,1} = w_{j,1}$, when $w_{ij,2}^{(k)} > w_{ji,2}^{(k)}$, object $j$ has greater influence to $i$ than that from $j$ to $i$.
%On the other hand, if $w_{ij,2}^{(k)} < w_{i,1}$, it denotes $i$ would more utilize its unary term's displacement estimation than $j$'s information.

A detailed derivation of Eq. (\ref{Equ:infer_param}) is given as follows. The mean-field method is to approximate the distribution $P({\bf d}|{\bf r}, I)$ with a distribution $Q({\bf d}|{\bf r}, I)$, which can be expressed as a product of independent marginals $Q({\bf d}|{\bf r}, I) = \prod_{1}^{N} Q_{i}( d_{i}|{\bf r}, I)$. The optimal approximation of $Q$ is obtained by minimizing Kullback-Leibler (KL) divergence between $P$ and $Q$. The solution for $Q$ has the following form,
\begin{equation}
  \log(Q_{i}( d_{i}|{\bf r}, I)) = E_{i \neq j}[\log(P({\bf d}|{\bf r}, I))] + \textnormal{const},
  \label{eqn:mf_theory}
\end{equation}
where $E_{i \neq j}$ denotes expectation under $Q$ distributions over all variables $d_{j}$ for $j \neq i$. The inference is formulated as
\begin{align}
  & \log(Q_{i}( d_{i}|{\bf r}, I)) =  \phi(d_i, r_i, I) + \sum_{i,j} \psi(d_i, d_j, r_i, r_j, I) \nonumber \\
     &  =  w_{i,1} (d_i - f_d(r_i, I))^2 \nonumber \\
     & \,\,\,\,\,\, + (1-w_{i,1}) \sum_{i \neq j} \sum_{k} w_{ij,2}^{(k)} (\Delta d_{ij} - \Delta s_{ij})^2 \label{Equ:infer_derivation} \\
     &  =  (w_{i,1}+ (1-w_{i,1})\sum_{i \neq j} \sum_{k} w_{ij,2}^{(k)}) d_i^2 \nonumber \\
     & \,\,\,\,\,\, - 2(w_{i,1}f_d(r_i, I)+ (1-w_{i,1}) \sum_{i \neq j} \sum_{k} w_{ij,2}^{(k)}(d_j+\Delta s_{ij})) d_i \nonumber \\
     & \,\,\,\,\,\, + \textnormal{const}. \nonumber
\end{align}
Each $\log(Q_{i}(d_{i}|{\bf r}, I))$ is a quadratic form with respect to $d_{i}$ and its means therefore are
\begin{align}
  \mu_{i} = \frac{w_{i,1}f_d(r_i, I)+ (1-w_{i,1})\sum_{i \neq j} \sum_{k} w_{ij,2}^{(k)}(d_j-\Delta s_{ij})}{w_{i,1}+(1-w_{i,1})\sum_{i \neq j} \sum_{k} w_{ij,2}^{(k)}}.
\end{align}
The inference task is to minimize $P({\bf d}|{\bf r}, I)$. Since we approximate conditional distribution with product of independent marginals, an estimate of each $d_{i}$ is obtained as the expected value $\mu_{i}$ of the corresponding quadratic function,
\begin{align}
  \widehat{d_{i}} = \arg\min_{d_{i}}(Q_{i}(d_{i}|{\bf r}, I)) = \mu_{i}.
\end{align}

\subsection{The Overall MOT Algorithm}
\label{ssec:overall}

The overall algorithm with our proposed DCCRF is shown in Algorithm \ref{alg:overall}. At each time $t$, the DCCRF inputs are existing tracklets up time $t-1$, and consecutive frames at time $t-1$ and time $t$. It outputs each tracklet's displacement estimation. After obtaining displacement estimations $\widehat{d_i}$ for each tracklet $r_i$ by DCCRF, its estimated location at time $t$ can be simply calculated as the summation of its location $b_{r_i}$ at time $t-1$ and its estimated displacement $\widehat{d_i}$, i.e.,
\begin{align}
\widehat{b_{r_i}} = b_{r_i} + \widehat{d_i}.
\label{eqn:prediction}
\end{align}
Based on such estimated locations, we utilize a visual-similarity CNN (Section \ref{sssec:assignment_cnn}) as well as the Intersection-over-Union value as the criterion for tracklet-detection association to generate longer tracklets (Section \ref{sssec:association}). To make our online MOT system complete, we also specify our detailed strategies for tracklet initialization (Section \ref{sssec:initialization}), occlusion handling and tracklet termination (Section \ref{sssec:occlusion}).

\begin{algorithm}
\label{alg:overall}
\caption{The overall MOT algorithm}
\LinesNumbered
\KwIn{Images sequence up to time $t$, per-frame object detections $b_1, b_2, b_3, \cdots$}
\KwOut{Object tracklets up to time $t$.}
\For{time = $1,\cdots,t$}{
	Estimate tracked object displacements $d_i$ (Section \ref{ssec:dccrf})\;
	Estimate tracklet location $\widehat{b_{r_i}}$ (Eq. (\ref{eqn:prediction}))\;
	Calculate tracklet-detection similarities $(\widehat{b_{r_i}}, b_j)$ (Section \ref{sssec:association})\;
	Hungarian algorithm	to obtain tracklet-associated detection $b_{r_i}$ (Section \ref{sssec:association})\;
	\For{each tracklet $r_i$}{
		\uIf{IoU$(\widehat{b_{r_i}}, b_j)\geq 0.5$}{Append $b_j$ to tracklet ${r_i}$\;}
		\uElseIf{$0.5 >$IoU$(\widehat{b_{r_i}}, b_j)\geq 0.3$}{Append $(b_j + \widehat{b_{r_i}}) / 2$ to tracklet ${r_i}$\;}
		\Else{
			$r_i$ has no detection association\;
			\eIf{no association $>m$ frames}{Tracklet termination (Section \ref{sssec:occlusion})\;}
			{Append $\widehat{b_{r_i}}$ to tracklet $r_i$ (Section \ref{sssec:occlusion})\;}
		}
	}
	\For{detections not associated to tracklets}{
		\If{high overall similarity for $k$ frames}{
			Tracklet initialization (Section \ref{sssec:initialization})\;
		}	
	}
}
\end{algorithm}

\subsubsection{Visual-similarity CNN}
\label{sssec:assignment_cnn}

The tracklet-detection associations need to be determined based on visual cues and spatial cues simultaneously. We propose a visual-similarity CNN for calculating visual similarities between image patches cropped at bounding box locations in the same frame. The visual-similarity CNN has similar network structure as our visual-displacement CNN in Section \ref{sssec:unary}. However, the network takes image patches in the same video frame as inputs and outputs the confidence whether the input pair represents the same object. It is therefore trained with a binary cross-entropy loss. In addition, the training samples are generated differently for the visual-similarity CNN. Instead of cropping two consecutive video frames at the same bounding box locations as the visual-displacement CNN, the visual-similarity CNN requires positive pairs to be cropped at different locations of the same object at anytime in the same video, while the negative pairs to be image patches belonging to different objects. For cropping image patches, we dont't enlarge the object's bounding box, which is also different to our visual-displacement CNN. During training, the ratio between positive and negative pairs are set to $1$:$3$ and the network is trained similarly to that of visual-displacement CNN. 

\subsubsection{Tracklet-detection association}
\label{sssec:association}

Given the estimated tracklet locations and detection boxes at time $t$, they are associated with detection boxes based on the visual and spatial similarities between them.
The associated detection boxes can then be appended to their corresponding tracklets to form longer ones up to time $t$.
%The associations between them are determined based on both their visual and spatial similarities.
Let $\widehat{b_{r_i}}$ and $b_j$ denote the $i$th tracklet's estimated location and the $j$th detection box at time $t$. Their visual similarity calculated by the visual-similarity CNN in Section \ref{sssec:assignment_cnn} is denoted as $V(\widehat{b_{r_i}}, b_j)$. The spatial similarity between the estimated tracklet locations and detection boxes are measured as the their box Intersection-over-Union values $IoU(\widehat{b_{r_i}}, b_j)$. If a detection box is tried to be associated with multiple tracklets, Hungrian algorithm is utilized to determine the optimal associations with the following overall similarity,
\begin{align}
S(\widehat{b_{r_i}}, b_j) = V(\widehat{b_{r_i}}, b_j) + \lambda IoU(\widehat{b_{r_i}}, b_j),
\label{eqn:overall_similarity}
\end{align}
where $\lambda$ is the weight balancing the visual and spatial similarities and is set to 1 in our experiments. After the box association by Hungarian algorithm, if a tracklet is associated with a detection box that has an IoU value greater than 0.5 with it, the associated detection box are directly appended to the end of the tracklet. If the IoU value is between 0.3 and 0.5, the average of the associated detection box and estimated tracklet box are appended to the tracklet to compensate for the possible noisy detection box. If the IoU value is smaller than 0.3, tracklet might be considered as being terminated or temporally occluded (Section \ref{sssec:occlusion}).

%The DCCRF outputs the candidate locations for each past tracklet at time $t+1$. The detection result that has the highest IoU value and appearance similarity score with the candidate box is associated to the tracklet to form a longer one. If an object detection result at time t is tried to be associated with multiple past tracklets, Hungrian algorithm is utilized to associate the detections with each tracklet's association confidences. The cost $c(i,j)$ is calculated as follows. The $P_{a}$ is the similarity confidence.
%\begin{equation}
%  c(i,j) = 2 - P_{a}(T_{i}^{t}, b_{j}) - IOU(T_{i}^{t}, b_{j})
%  \label{Equ:cost}
%\end{equation}

\subsubsection{Tracklet initialization}
\label{sssec:initialization}

If an object detection box at time $t-1$ is not associated to any tracklet in the above tracklet-detection association step, it is treated as a \emph{candidate box} for initializing new tracklets. For each such candidate box at time $t-1$, its visually inferred displacement between time $t-1$ and $t$ is first  obtained by our visual-displacement CNN in Section \ref{sssec:unary}. Its estimated box location can be easily calculated following Eq. (\ref{eqn:prediction}). The visual similarities $V$ and spatial similarities $IoU$ between the estimated box at $t$ and candidate boxes at $t$ are calculated. To form new candidate tracklet, the candidate box at time $t-1$ is only associated with the candidate box at time $t$ that has 1) greater-than-0.3 $IoU$ and 2) greater-than-0.8 visual similarity with its estimated box location.
If there are multiple candidate associations, Hungarian algorithm is utilized to associate the candidate box at $t$ to its optimal candidate association at $t-1$ according to the overall similarities (Eq. (\ref{eqn:overall_similarity})). If none of the candidate associations at time $t$ satisfies the above two conditions with the candidate box at $t-1$, the candidate box is ignored and would not be used for tracking initialization.
Such operations are iterated over time to generate longer candidate tracklets. If a candidate tracklet is over $k$ frames ($k = 4$ for pedestrain tracking with 25-fps videos), it is initialized as a new tracklet.

\subsubsection{Occlusion handling and tracklet termination}
\label{sssec:occlusion}

If a past tracklet is not associated to any detection box at time $t$, the tracked object is considered as being possibly occluded or temporally missed. For a possibly occluded object, we directly associate its past tracklet to its estimated location by our DCCRF at time $t$ to create a virtual tracklet. The same operation is iterated for $m$ frames, i.e., if the virtual tracklet is not associated to any detection box  for more than $m$ time steps, the virtual tracklet is terminated. For pedestrian tracking, we empirically set $m = 5$.

\section{Experiments}
\label{sec:experiments}

In this section, we present experimental results of the proposed online MOT algorithm. We first introduce evaluation datasets and implementation details for our proposed framework in Sections \ref{ssec:datasets} and \ref{ssec:details}. In Section \ref{ssec:comparison}, we compare the proposed method with state-of-the-art approaches on the public MOT datasets. The individual components of our proposed method are evaluated in Section \ref{ssec:components}.

\subsection{Datasets and Evaluation Metric}
\label{ssec:datasets}

We conduct experiments on the 2DMOT15 \cite{leal2015motchallenge} and 2DMOT16 \cite{milan2016mot16} benchmarks, which are widely used to evaluate the performance of MOT methods. Both of them have two tracks: public detection boxes \cite{xiang2015learning,hong2016online,bae2017confidence} and private detection boxes \cite{Henschel2017ImprovementsTF,Yu2016POI}. For comparing with only the performance of tracking algorithms, we evaluate our method with the provided public detection boxes.

\subsubsection{2DMOT15}

This dataset is one of the largest datasets with moving or static cameras, different viewpoints and different weather conditions. It contains a total of 22 sequences, half for training and half for testing, with a total of 11286 frames (or 996 seconds). The training sequences contain over 5500 frames, 500 annotated trajectories and 39905 annotated bounding boxes. The testing sequences contain over 5700 frames, 721 annotated trajectories and 61440 annotated bounding boxes.
%It is diffcult for algorithms to overfit such a large amount of data.
The public detection boxes in  2DMOT15 are generated with aggregated channel features (ACF).

\subsubsection{2DMOT16}

This dataset is an extension to 2DMOT15. Compared to 2DMOT15, new sequences are added and the dataset contains almost 3 times more bounding boxes for training and testing. Most sequences are in high resolution, and the average pedestrian number in each video frame is 3 times higher than that of the 2DMOT15. In 2DMOT16, deformable part models (DPM) based methods are used to generate public detection boxes, which are more accurate than boxes in 2DMOT15.

\subsubsection{Evaluation Metric}

For the quantitative evaluation, we adopt the popular CLEAR MOT metrics \cite{leal2015motchallenge}, which include:
\begin{itemize}
  \item \textbf{MOTA:} Multiple Object Tracking Accuracy. This metric is usually chosen as the main performance indicator for MOT methods. It combines three types of errors: false positives, false negatives, and identity switches.
  \item \textbf{MOTP:} Multiple Object Tracking Precision. The misalignment between the annotated and the predicted bounding boxes.
  \item \textbf{MT:} Mostly Tracked targets. The ratio of ground-truth trajectories that are covered by a track hypothesis for at least 80\% of their respective life span.
  \item \textbf{ML:} Mostly Lost targets. The ratio of ground-truth trajectories that are covered by a track hypothesis for at most 20\% of their respective life span.
  \item \textbf{FP:} The total number of false positives.
  \item \textbf{FN:} The total number of false negatives (missed targets).
  \item \textbf{ID Sw:} The total number of identity switches. Please note that we follow the stricter definition of identity switches as described in MOT challenge.
  \item \textbf{Frag:} The total number of times a trajectory is fragmented (i.e., interrupted during tracking).
\end{itemize}
%The most important performance indicator is MOTA, which combines FP, FN and ID Sw.

\subsection{Implementation details}
\label{ssec:details}

\subsubsection{Training schemes and setting}

%We train our proposed DCCRF in three stages. In the first stage, the proposed visual-displacement CNN is trained with a cross-entropy loss. The CNN is trained with Stochastic Gradient Descent (SGD) with a batch size of XXX and a learning rate of XXX. The learning rate is decreased to $1/10$ of its current value after every five epochs and the CNN is trained with 15 epochs. We randomly sample image patches at detection boxes from all training frames. In each mini-batch, the ratio between positive (matched) and negative (unmatched) image pairs is set to $1:3$. In the second stage, the visual-displacement CNN is fixed and other learnable parameters in our DCCRF is trained with randomly sampled pairs of consecutive frames as training samples. The $L_1$ loss is adopted as supervision signals,
%\begin{equation}
%  \zeta_{loss} = \sum \| \widehat{d_{i}} - d_i^{gt} \|_{1},
%  \label{eqn:l1loss}
%\end{equation}
%where $\widehat{d_{i}}$ and $d_i^{gt}$ are estimated displacements and the ground-truth displacements. In the third stage, the overall DCCRF is trained in an end-to-end manner.

For visual-displacement and visual-similarity CNNs, we adopt ResNet-101 \cite{He_2016_CVPR, zeng2016crafting} as the network structure and replace the topmost layer to output displacement confidence or same-object confidence. Both CNN are pretrained on the ImageNet dataset. For cropping image patches from $\bar{b}_i$, we enlarge each detection box $b_i$ by a factor of 5 in width and 2 in height to obtain $\bar{b}_i$. Image patches for the two CNNs are cropped at the same locations from consecutive frames as described in Section \ref{sssec:unary}, which are then resized to $224 \times 224$ as the CNN inputs.

We train our proposed DCCRF in three stages. In the first stage, the proposed visual-displacement CNN is trained with the cross-entropy loss and batch Stochastic Gradient Descent (SGD) with a batch size of 5. The initial learning rate is set to $10^{-6}$ and is decreased by a factor of 1/10 every 50,000 iterations. The training generally converges after 600,000 iterations. In the second stage, the learned visual-displacement CNN from stage-1 is fixed and other parameters in our DCCRF are trained with $L_1$ loss,
\begin{equation}
  \zeta_{loss} = \sum \| \widehat{d_{i}} - d_i^{gt} \|_{1},
  \label{eqn:l1loss}
\end{equation}
where $\widehat{d_{i}}$ and $d_i^{gt}$ are estimated displacements and the ground-truth displacements for the $i$th tracked object.
In the final stage, the DCCRF is trained in an end-to-end manner with the above $L_1$ loss and the cross-entropy loss for visual-displacement CNN in unary terms. We find that 5 iterations of the mean-field approximation generate satisfactory results. The DCCRF is trained with an initial learning rate of $10^{-4}$, which is decreased by a factor of 1/3 every 5,000 iterations. The training typically converges after 3 epochs.

Our code is implemented with MATLAB and Caffe. The overall tracking speed of the proposed method on MOT16 test sequences is 0.1 fps using the 2.4GHz CPU and a Maxwell TITAN X GPU without some acceleration library packages.

%For detection boxes, we enlarge the box by a factor of 5 in width and 2 in height. Image patches are cropped at the same locations of consecutive frames as described in Section \ref{sssec:unary}, which are then resized to $224 \times 224$ as inputs. The visual-displacement CNN is trained with the cross-entropy loss and batch Stochastic Gradient Descent (SGD) with a batch size of 5. The initial learning rate is $1e-6$ and is decreased by a factor of 10 every 50,000 iterations. The training is converged after 600,000 iterations. For the training of CRF, the parameters are learned by back-propagation algorithm when the motion CNN parameters are fixed. After obtaining the first stage's parameters, we use a small learning rate $xxx$ to tune the unified network. For end-to end training, the learning rate is set to be lower (0.0001) and is dropped every 5k iterations also by a factor of 3. The unified end-to-end model typically converges after 3 epochs. The CCRF layer takes the objects' displacements as inputs. Please note that setting the proper initial parameters in CCRF layer optimization stage can help accelerate and stabilize the training process. We find that five iterations of approximate inference already provides satisfactory results in our experiments. The loopy structure used in this implementation is visualized in Fig. (\ref{Fig:OverallFramework}). Spatially, the structured model has edges connecting with each person.

\subsubsection{Data augmentation}

To introduce more variation into the training data and thus reduce possible overfitting, we augment the training data. For pre-training the visual-displacement CNN, the input images are image patches centered at detection boxes. We augment the training samples by random flipping as well as randomly shifting the cropping positions by no more than $\pm 1/5$ of detection box width or height for $x$ and $y$ dimensions respectively. For end-to-end training the DCCRF, except for random flipping of whole video frames, the time intervals between the two input video frames are randomly sampled from the interval of $[1,3]$ to generate more frame pairs with larger possible displacements between them.

\begin{table*}[tb]
\caption{Quantitative results by our method and state-of-the-art MOT methods on 2DMOT15 dataset. Bold numbers indicate the best results of online or offline methods respectively). $\uparrow$ denotes that higher is better and $\downarrow$ represents the opposite.}
\label{table_mot15}
\centering
\begin{tabular}{|c|c|c|c|c|c|c|c|c|c|c|}
\hline
Tracking Mode & Method & \textbf{MOTA}$\uparrow$ & MOTP$\uparrow$ & MT$\uparrow$ & ML$\downarrow$ & FP$\downarrow$ & FN$\downarrow$ & ID Sw$\downarrow$ & Frag$\downarrow$\\
\hline
\hline
Offline & SMOT \cite{dicle2013way} & 18.2\% & 71.2\% & 2.8\% & 54.8\% & 8780 & 40310 & 1148 & 2132\\
\hline
Offline & CEM \cite{milan2014continuous} & 19.3\% & 70.7\% & 8.5\% & 46.5\% & 14180 & 34591 & 813 & 1023\\
\hline
Offline & DCO\_X \cite{milan2016multi} & 19.6\% & 71.4\% & 5.1\% & 54.9\% & 10652 & 38232 & 521 & {\textbf{819}} \\
\hline
Offline & SiameseCNN \cite{leal2016learning} & 29.0\% & 71.2\% & 8.5\% & 48.4\% &\textbf{5160} & 37798 & 639 & 1316\\
\hline
Offline & CNNTCM \cite{wang2016joint} & 29.6\% & 71.8\% & 11.2\% & \textbf{44.0\%} & 7786 & 34733 & 712 & 943\\
\hline
Offline & NOMT \cite{choi2015near} & \textbf{33.7\%} & \textbf{71.9\%} & \textbf{12.2\%} & \textbf{44.0\%} & 7762 & \textbf{32547} & \textbf{442} & 823\\
\hline
\hline
Online & TC\_ODAL \cite{bae2014robust} & 15.1\% & 70.5\% & 3.2\% & 55.8\% & 12970 & 38538 & 637 & 1716\\
\hline
Online & RNN\_LSTM \cite{milan2017online} & 19.0\% & 71.0\% & 5.5\% & 45.6\% & 11578 & 36706 & 1490 & 2081\\
\hline
Online & RMOT \cite{yoon2015bayesian} & 18.6\% & 69.6\% & 5.3\% & 53.3\% & 12473 & 36835 & 684 & 1282\\
\hline
Online & oICF \cite{kieritz2016online} & 27.1\% & 70.0\% & 6.4\% & 48.7\% & 7594 & 36757 & \textbf{454} & 1660\\
\hline
Online & SCEA \cite{hong2016online} & 29.1\% & 71.1\% & 8.9\% & 47.3\% & 6060 & 36912 & 604 & \textbf{1182}\\
\hline
Online & MDP \cite{xiang2015learning} & 30.3\% & \textbf{71.3\%} & \textbf{13.0\%} & 38.4\% & 9717 & \textbf{32422} & 680 & 1500 \\
\hline
Online & CDA\_DDAL \cite{bae2017confidence} & 32.8\% & 70.7\% & 9.7\% & 42.2\% & \textbf{4983} & 35690 & 614 & 1583 \\
\hline
Online & Proposed Method & \textbf{33.6\%} & 70.9\% & 10.4\% & \textbf{37.6\%} & 5917 & 34002 & 866 & 1566\\
\hline
\end{tabular}
\end{table*}

\begin{table*}[tb]
\caption{Quantitative results by our proposed method and state-of-the-art MOT methods on 2DMOT16 dataset. $\uparrow$ denotes that higher is better and $\downarrow$ represents the opposite.}
\label{table_mot16}
\centering
\begin{tabular}{|c|c|c|c|c|c|c|c|c|c|c|}
\hline
Tracking Mode & Method & \textbf{MOTA}$\uparrow$ & MOTP$\uparrow$ & MT$\uparrow$ & ML$\downarrow$ & FP$\downarrow$ & FN$\downarrow$ & ID Sw$\downarrow$ & Frag$\downarrow$\\
\hline
\hline
Offline & TBD \cite{geiger20143d} & 33.7\% & 76.5\% & 7.2\% & 54.2\% & 5804 & 112587 & 2418 & 2252 \\
\hline
Offline & LTTSC-CRF \cite{le2016long} & 37.6\% & 75.9\% & 9.6\% & 55.2\% & 11969 & 101343 & 481 & 1012 \\
\hline
Offline & LINF \cite{fagot2016improving} & 41.0\% & 74.8\% & 11.6\% & 51.3\% & 7896 & 99224 & 430 & 963\\
\hline
Offline & MHT\_DAM \cite{kim2015multiple} & 42.9\% & \textbf{76.6\%} & 13.6\% & 46.9\% & \textbf{5668} & 97919 & 499 & 659\\
\hline
Offline & JMC \cite{tang2016multi} & 46.3\% & 75.7\% & 15.5\% & \textbf{39.7\%} & 6373 & 90914 & 657 & 1114 \\
\hline
Offline & NOMT \cite{choi2015near} & \textbf{46.4\%} & \textbf{76.6\%} & \textbf{18.3\%} & 41.4\% & 9753 & \textbf{87565} & \textbf{359} & \textbf{504}\\
\hline
\hline
Online & OVBT \cite{ban2016tracking} & 38.4\% & 75.4\% & 7.5\% & 47.3\% & 11517 & 99463 & 1321 & 2140 \\
\hline
Online & EAMTT\_pub \cite{sanchez2016online} & 38.8\% & 75.1\% & 7.9\% & 49.1\% & 8114 & 102452 & 965 & 1657 \\
\hline
Online & oICF \cite{kieritz2016online} & 43.2\% & 74.3\% & 11.3\% & 48.5\% & 6651 & 96515 & \textbf{381} & 1404 \\
\hline
Online & CDA\_DDAL \cite{bae2017confidence} & 43.9\% & 74.7\% & 10.7\% & 44.4\% & 6450 & 95175 & 676 & 1795  \\
\hline
Online & Proposed Method & \textbf{44.8\%} & \textbf{75.6\%} & \textbf{14.1\%} & \textbf{42.3\%} & \textbf{5613} & \textbf{94125} & 968 & \textbf{1378} \\
\hline
\end{tabular}
\end{table*}

\begin{figure*}[t]
 \centering
 \begin{tabular}{c@{\hspace{-2mm}}c@{\hspace{2mm}}c@{\hspace{2mm}}c}
  &
        \includegraphics[width=5cm]{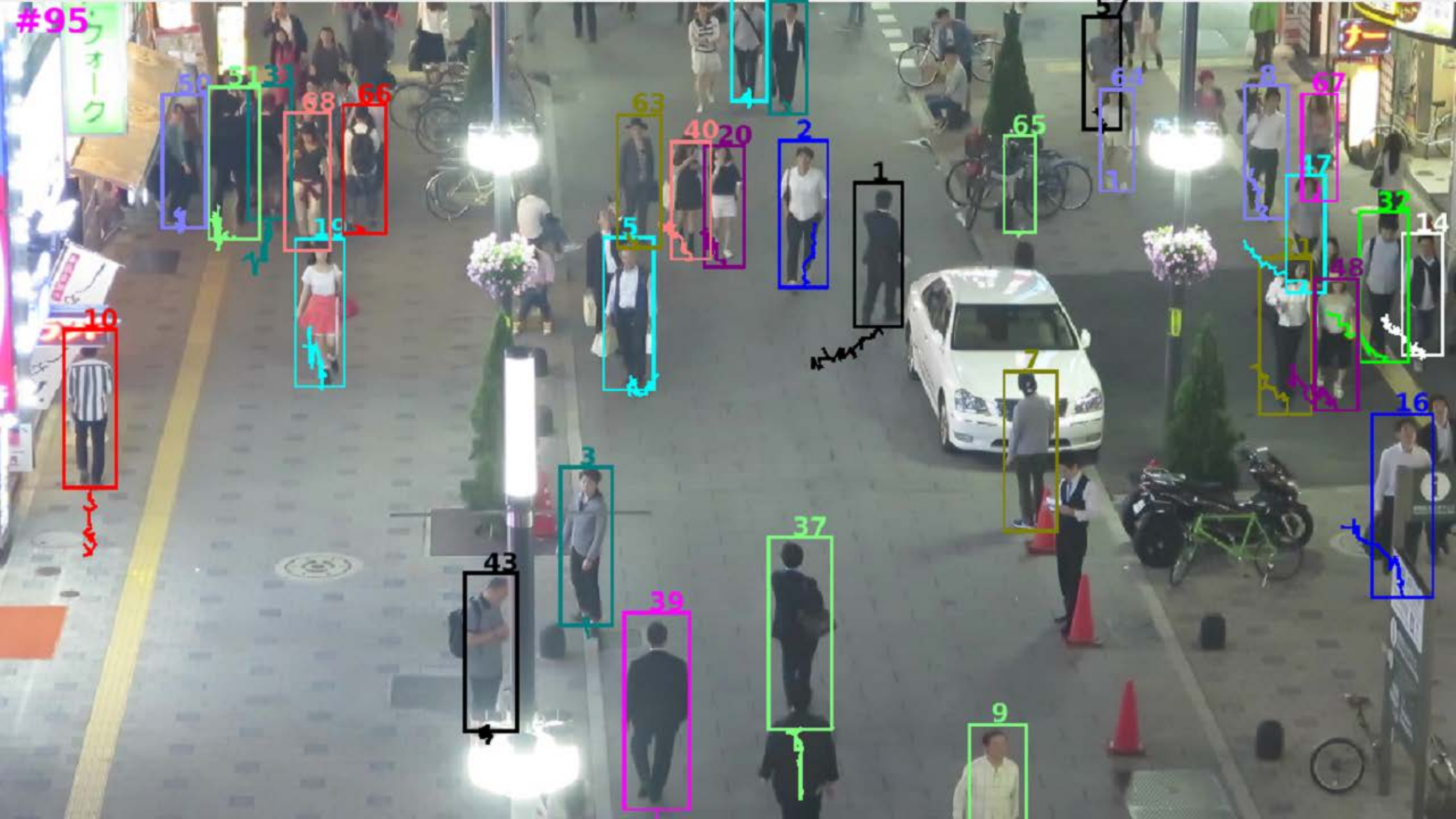}&
        \includegraphics[width=5cm]{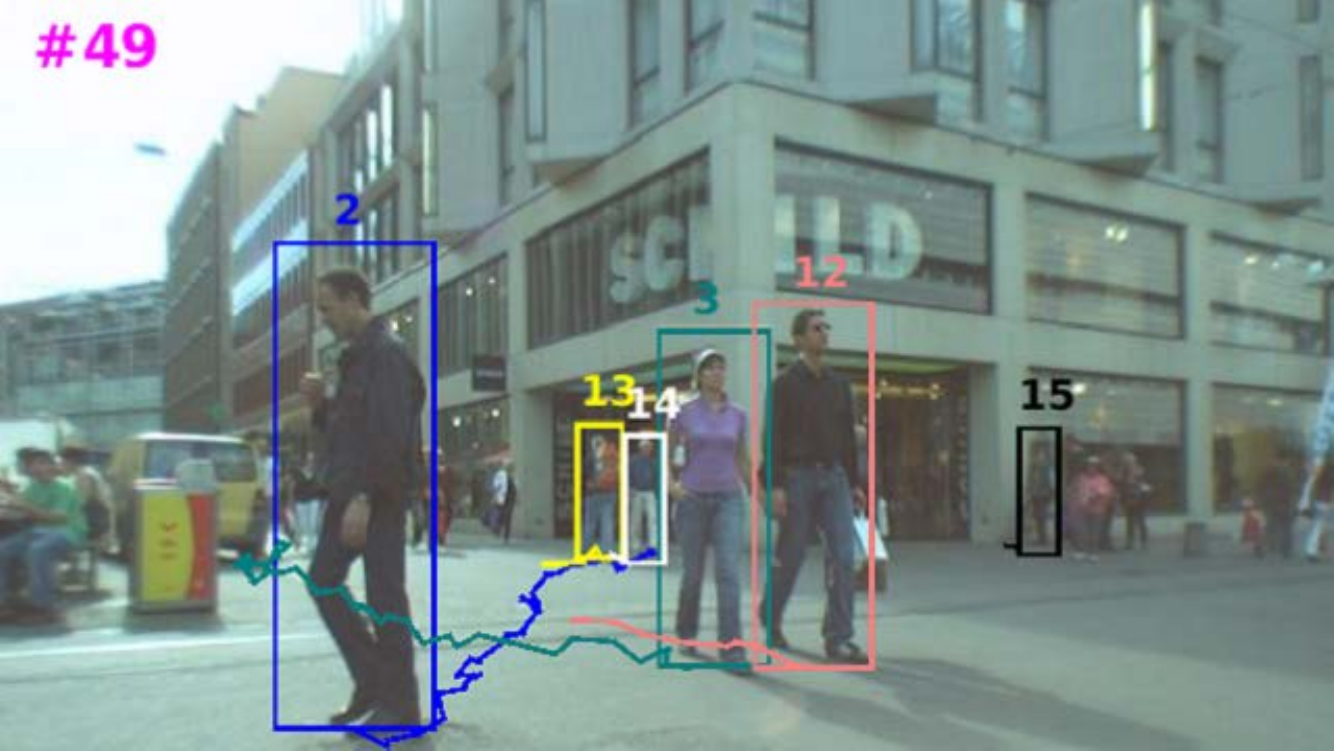}&
  \includegraphics[width=5cm]{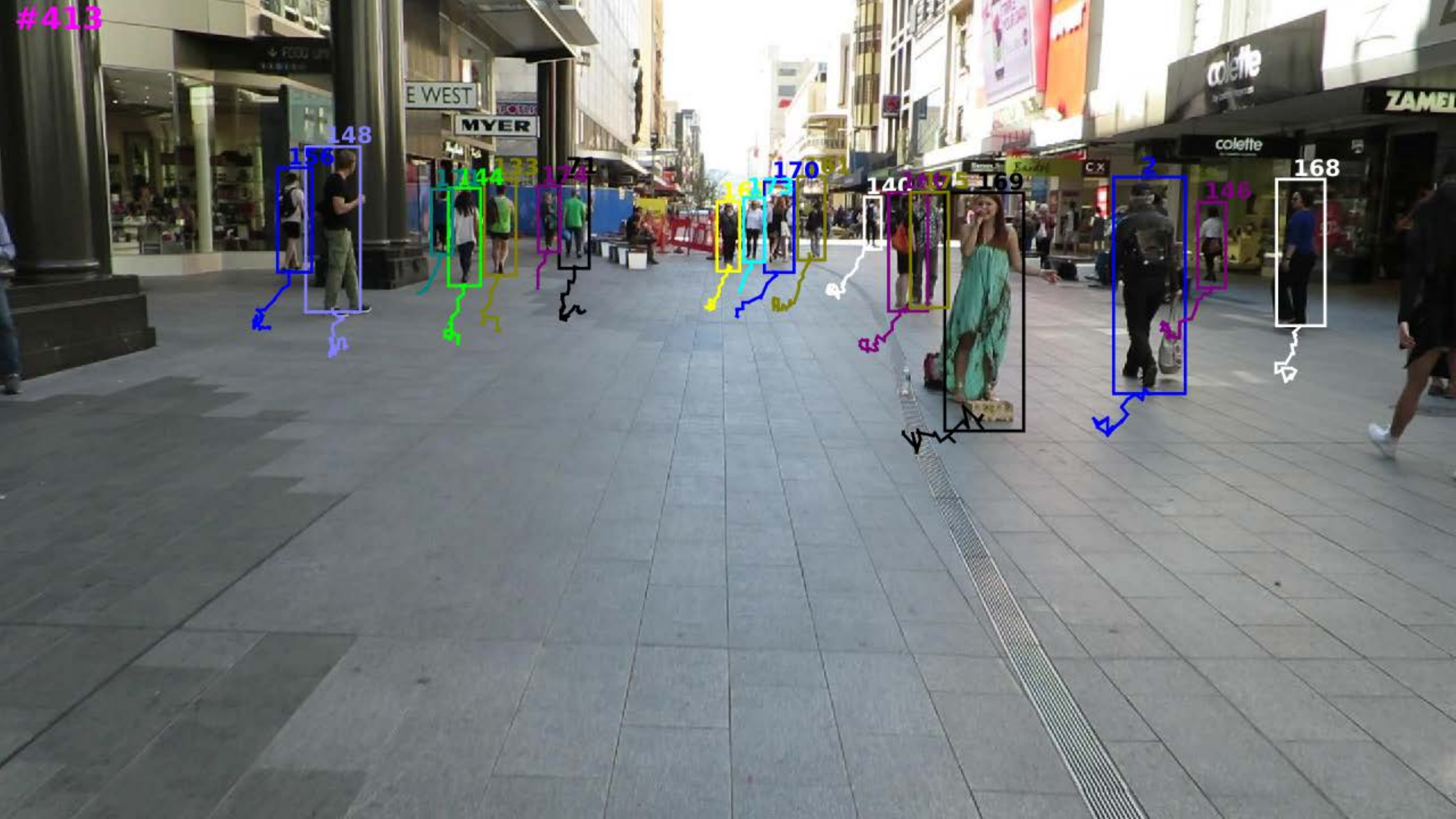}\\
         & (a) MOT1603 & (b) MOT1606 & (c) MOT1607 \\
 \end{tabular}
    \begin{tabular}{c@{\hspace{-2mm}}c@{\hspace{2mm}}c@{\hspace{2mm}}c}
  &
  \includegraphics[width=5cm]{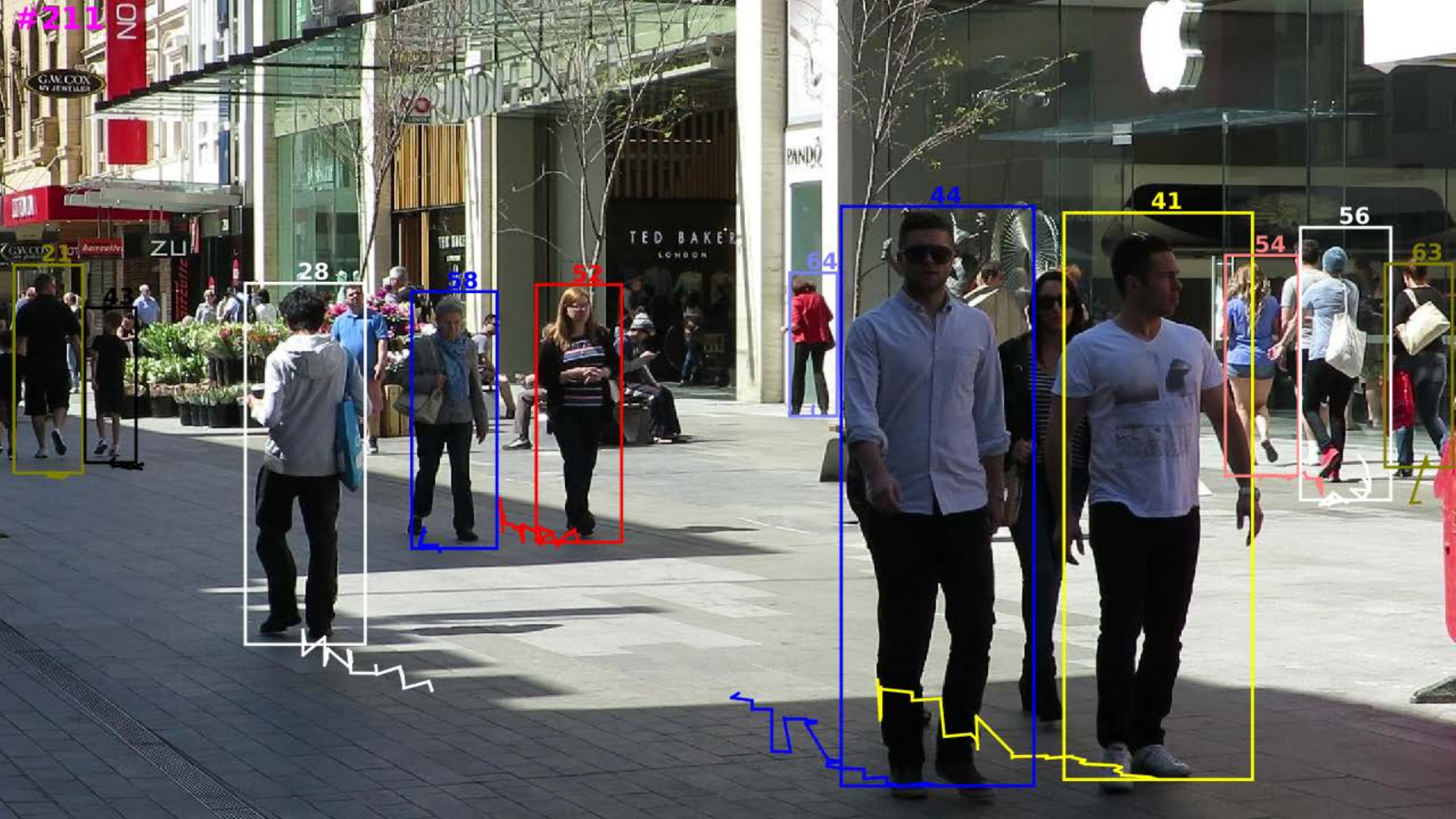}&
        \includegraphics[width=5cm]{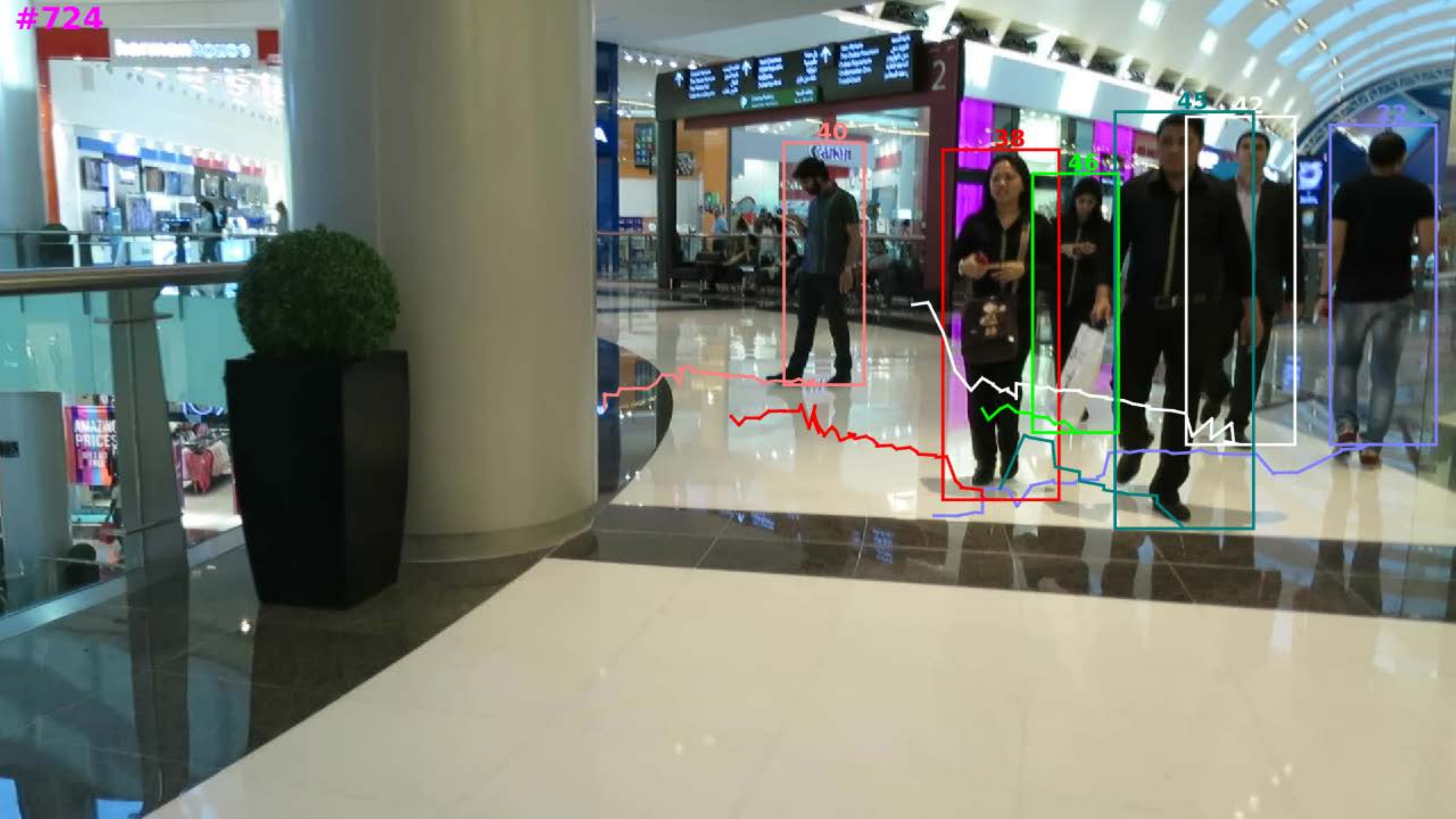}&
  \includegraphics[width=5cm]{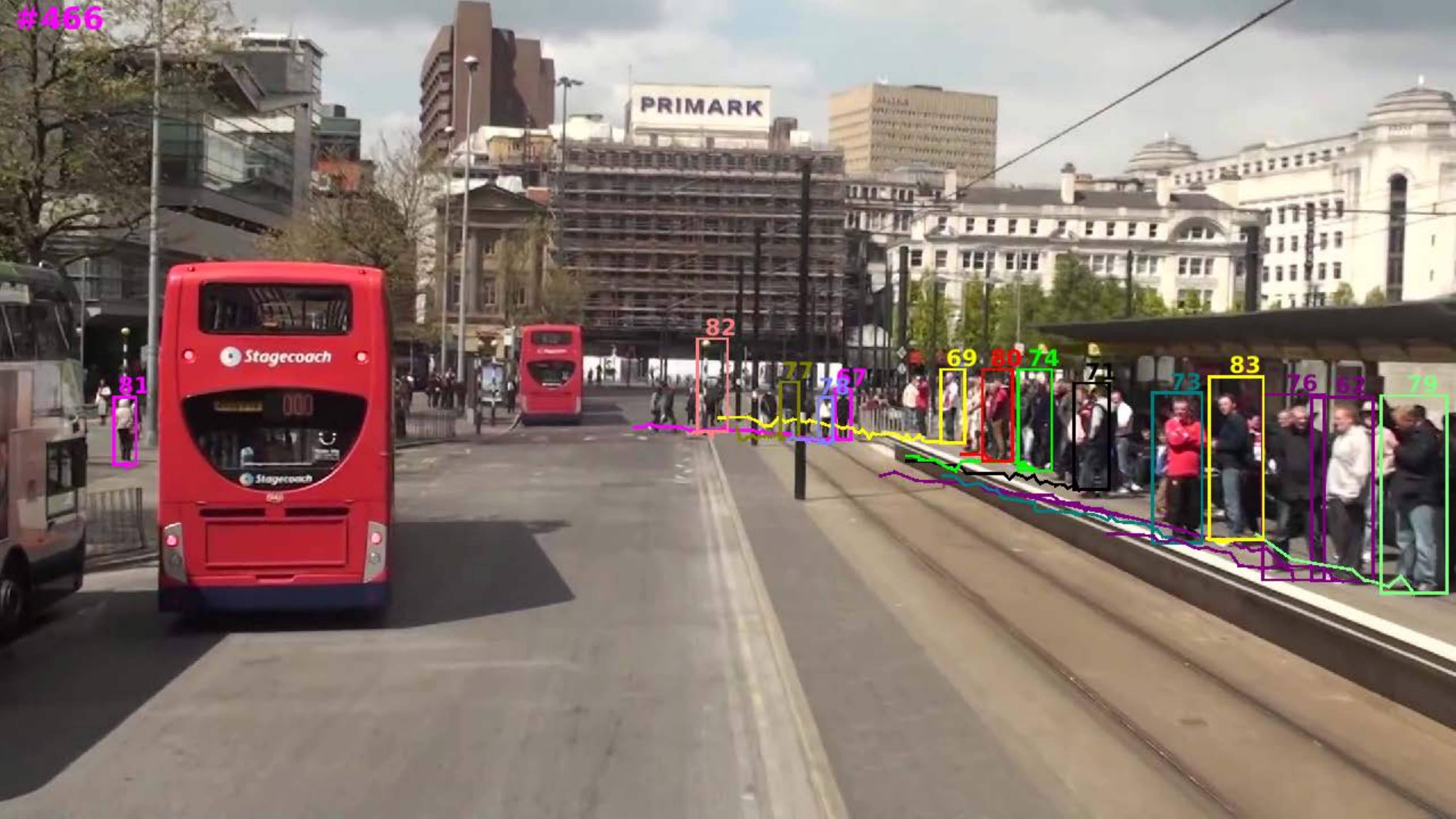}\\
        & (d) MOT1608 & (e) MOT1612 & (f) MOT1614 \\
 \end{tabular}
  \caption{Example tracking results by our proposed method on 2DMOT16 dataset.}
  \label{fig:results}
\end{figure*}

\subsection{Quantitative results on 2DMOT15 and 2DMOT16}
\label{ssec:comparison}

On the MOT2015 and MOT2016 datasets, we test our proposed method and compare it with state-of-the-art MOT methods\footnote{Note that only methods in peer-reviewed publications are compared in this paper. ArXiv papers that have not undergone peer-review are not included.} including SMOT \cite{dicle2013way}, MDP \cite{xiang2015learning}, SCEA \cite{hong2016online}, CEM \cite{milan2014continuous}, RNN\_LSTM \cite{milan2017online}, RMOT \cite{yoon2015bayesian}, TC\_ODAL \cite{bae2014robust}, CNNTCM \cite{wang2016joint}, SiameseCNN \cite{leal2016learning}, oICF \cite{kieritz2016online}, NOMT \cite{choi2015near}, CDA\_DDAL \cite{bae2017confidence}. The results of the compared methods are listed in Tables \ref{table_mot15} and \ref{table_mot16}. We focus on the MOTA value as the main performance indicator, which is a weighted combination of false negatives (FN), false positives (FP) and identity switches (ID Sw).  Note that offline methods generally have higher MOTA than online methods because they can utilize not only past but also future information for object tracking and are only listed for reference here. Our proposed online MOT method outperforms all compared online methods and most offline methods \cite{xiang2015learning, hong2016online, kieritz2016online, bae2017confidence, leal2016learning}. As shown by the quantitative results, our proposed method is able to alleviate the difficulties caused by object mis-detections, noisy detections, and short-term occlusion. The qualitative results are shown in Fig. \ref{fig:results}.

Compared with SCEA \cite{hong2016online}, which also models inter-object interactions and speed differences to handle mis-detections caused by global camera motion, our learned DCCRF shows better performance, especially in FN for our more accurate displacement prediction which is able to recover more mis-detections. Our proposed method also outperforms MDP \cite{xiang2015learning} in terms of MOTA and FP by a large margin. MDP learns to predict four target states (active, tracked, lost and inactive) for each tracked object. However, it only models tracked object's movement patterns with a constant speed assumption, which is likely to result in false tracklet-detection associations and thus increases FP. CDA\_DDAL \cite{bae2017confidence} focuses on using discriminative visual features by a siamese CNN for tracklet-detection associations, which is not robust for occlusions and is easy to increase FN. Compared with other algorithms DCO\_X \cite{milan2016multi} and LTTSC-CRF \cite{le2016long} which also use conditional random field approximation to solve MOT problems, the results show that our proposed DCCRF has great advantages over other CRF-based methods in MOTA.

However, our method produces more ID switches than some compared methods, which is due to long-term occlusions that cannot be solved by our method.
 
\begin{table}[tb]
\caption{Component analysis of our proposed DCCRF on 2DMOT2016 dataset. $\uparrow$ denotes that higher is better and $\downarrow$ represents the opposite.}
\label{tab:component}
\centering
\begin{tabular}{|c|c|c|c|c|}
\hline
Method & \textbf{MOTA}$\uparrow$  & FP$\downarrow$ & FN$\downarrow$ & ID Sw$\downarrow$ \\
\hline
\hline
Proposed DCCRF & 44.8\% & 5613 & 94125 & 968 \\
\hline
\hline
Unary-only & 41.9\% & 7392 & 97618 & 876 \\
Unary-only+$L_1$-loss (reg) & 34.2\% & 12089 & 104810 & 3134  \\
\hline
\hline
DCCRF w/o size-asym & 43.6\% & 8063 & 93724 & 1035  \\
DCCRF w/o cfd-asym & 43.8\% & 7353 & 94163 & 969  \\
DCCRF w/ symmetry & 43.4\% & 9100 & 93076 & 1104 \\
\hline
\end{tabular}
\end{table}

\subsection{Component analysis on 2DMOT16}
\label{ssec:components}

To analyze the effectiveness of different components in our proposed framework, we also design a series of baseline methods for comparison.
The results of these baselines and our final method are reported in Table \ref{tab:component}. Similar to the above experiments, we focus on MOTA value as the main performance indicator. 1) Unary-only: this baseline utilizes only our unary terms in DCCRF, i.e., the visual-displacement CNN, with our overall MOT algorithm. Such a baseline model considers only tracked objects' appearance information. Compared with our proposed DCCRF, it has a $3\%$ MOTA drop, which denotes that the inter-object relations are crucial for regularizing each object's estimated displacement and should not be ignored. 2) Unary-only+$L_1$-loss (reg): since our visual-displacement CNN is trained with proposed cross-entropy loss instead of conventional $L_1$ or $L_2$ losses for regression problems, we train a visual-displacement CNN with smooth $L_1$-loss and test it in the same way as the above unary-only baseline. Compared with unary-only baseline, unary-only+$L_1$-loss has a significant $7\%$ MOTA drop, which demonstrates that our proposed cross-entropy loss results in much better displacement estimation accuracy. 3) DCCRF w/o cfd-asym and DCCRF w/o size-asym: the weighting functions of the pairwise term in our proposed DCCRF have two terms, a confidence-asymmetric term and a size-asymmetric term. We test using only one of them in our DCCRF's pairwise terms. The results show more than $1\%$ drop in terms of MOTA for both baseline methods compared with our proposed DCCRF, which validates the need of both terms in the weighting functions. 4) DCCRF w/ symmetry: this baseline method replaces the asymmetric pairwise term in our DCCRF with a symmetric one,
\begin{align}
\label{eq:gaussian_func}
(1-w_{i,1}) \sum_k \exp \left( -\frac{(l_i - l_j)^2}{2a_2^{(k)2}} \right)  (\Delta d_{ij} - \Delta s_{ij})^2,
\end{align}
where $l_i$ is the coordinates of $i$th object's center position and $a_2^{(k)}$ are learnable Gaussian kernel bandwidth parameters. Such a symmetric term assumes that the speed differences between close-by objects should be better maintained across time, while those between far-away objects are less regularized. There is a $1\%$ MOTA drop compared with our proposed DCCRF, which shows our asymmetric term is beneficial for the final performance. We also try to directly replace the sigmoid function in Eq. (\ref{eq:pairwise}) with a Gaussian-like function in the weighting function (Eq. (\ref{eq:gaussian_func})), which results in even worse performance.

\begin{table}[tb]
\begin{center}
\caption{Effects of different $\lambda$ parameter.}
\begin{tabular} {c|c c c}\hline
$\lambda$ & $0.5$ & $1$		& $1.5$ \\\hline
MOTA & $ 43.8\% $	& $ 44.8\% $ & $ 43.5\% $ \\\hline
\end{tabular}
\label{tab:balance_weight}
\end{center}
\end{table}
\begin{table}[tb]
\begin{center}
\caption{Results by different tracklet initialization parameter $k$.}
\begin{tabular} {c|c c c}\hline
$k$ & MOTA & FP	& FN \\\hline
$4$ & $44.8\%$	& $5613$ & $94125$ \\\hline
$8$ & $43.0\%$	& $4837$ & $98433$ \\\hline
\end{tabular}
\label{tab:k}
\end{center}
\end{table}
\begin{table}
\begin{center}
\caption{Results by different tracklet termination parameter $m$.}
\begin{tabular} {c|c c c}\hline
$m$ & MOTA & FP	& FN \\\hline
$5$ & $44.8\%$	& $5613$ & $94125$ \\\hline
$8$ & $44.7\%$	& $6861$ & $92976$ \\\hline
\end{tabular}
\label{tab:m}
\end{center}
\end{table}

In addition to the above, we also conduct experiments to analysize the effects of different hyper-parameters to show our DCCRF robustness. 1) The $\lambda$ controls the weight between the visual-similarity term and the DCCRF location prediction term for tracklet-detection association in Eq. (\ref{eqn:overall_similarity}). We test three different values of $\lambda$ and the results of different $\lambda$ are reported in Table \ref{tab:balance_weight}, which the final performance is not sensitive to the $\lambda$ value. 2) The $k$ is the length of a candidate tracklet to create an actual tracklet in section \ref{sssec:initialization}. We additionally test $k=8$ in Table \ref{tab:k}, which shows slightly performance drop, because larger $k$ will cause more low-confidence detections to be ignored. 3) The $m$ denotes the number of consecutive frames of missing objects to terminate its associated tracklet in section \ref{sssec:occlusion}. We additionally test $m = 8$ and the results in Table \ref{tab:m} show the peformance is not sensitive to the choice of $m$.

\section{Conclusion}
In this paper, we present the Deep Continuous Conditional Random Field (DCCRF) model with asymmetric inter-object constraints for solving the MOT problem. The unary terms are modeled as a visual-displacement CNN that estimates object displacements across time with visual information. The asymmetric pairwise terms regularize inter-object speed differences across time with both size-based and confidence-based weighting functions to weight more on high-confidence tracklets to correct tracking errors. By jointly training the two terms in DCCRF, the relations between objects' individual movement patterns and complex inter-object constraints can be better modeled and regularized to achieve more accurate tracking performance. Extensive experiments demonstrate the effectiveness of our proposed MOT framework as well as the individual components of our DCCRF.

%\textbf{Acknowledgment}: This work is supported in part by SenseTime Group Limited, in part by the General Research Fund through the Research Grants Council of Hong Kong under Grants CUHK14213616, CUHK14206114, CUHK14205615, CUHK419412, CUHK14203015, CUHK14239816, CUHK14207814, CUHK14208417, CUHK14202217, in part by the Hong Kong Innovation and Technology Support Programme Grant ITS/121/15FX, in part by the National Natural Science Foundation of China under Grant 61671125, Grant 61201271, and Grant 61301269, and in part by the China Postdoctoral Science Foundation under Grant 2014M552339.

% or


\ifCLASSOPTIONcaptionsoff
  \newpage
\fi



\bibliographystyle{IEEEtran}
%\bibliography{ref}

{\small
\bibliographystyle{ieee}

\end{document}